%% file: main.tex
\documentclass[letterpaper]{article} 
\usepackage[preprint]{aaai2027}  
\usepackage[hyphens]{url}  
\usepackage{graphicx} 
\usepackage{amsmath}
\usepackage{amsfonts}
\usepackage{natbib}  
\usepackage{caption} 
\usepackage{algorithm}
\usepackage{algorithmic}
\usepackage{booktabs}
\usepackage{colortbl}
\newcommand{\methodname}{EMPIRE}
\newcommand{\tok}[1]{\texttt{\textless #1\textgreater}}
\newcommand{\planablationref}{the supplementary analysis of Stage~II training with predicted versus ground-truth plans}
\newcommand{\methodlogo}{%
  \raisebox{-0.25em}{\includegraphics[height=1.45em]{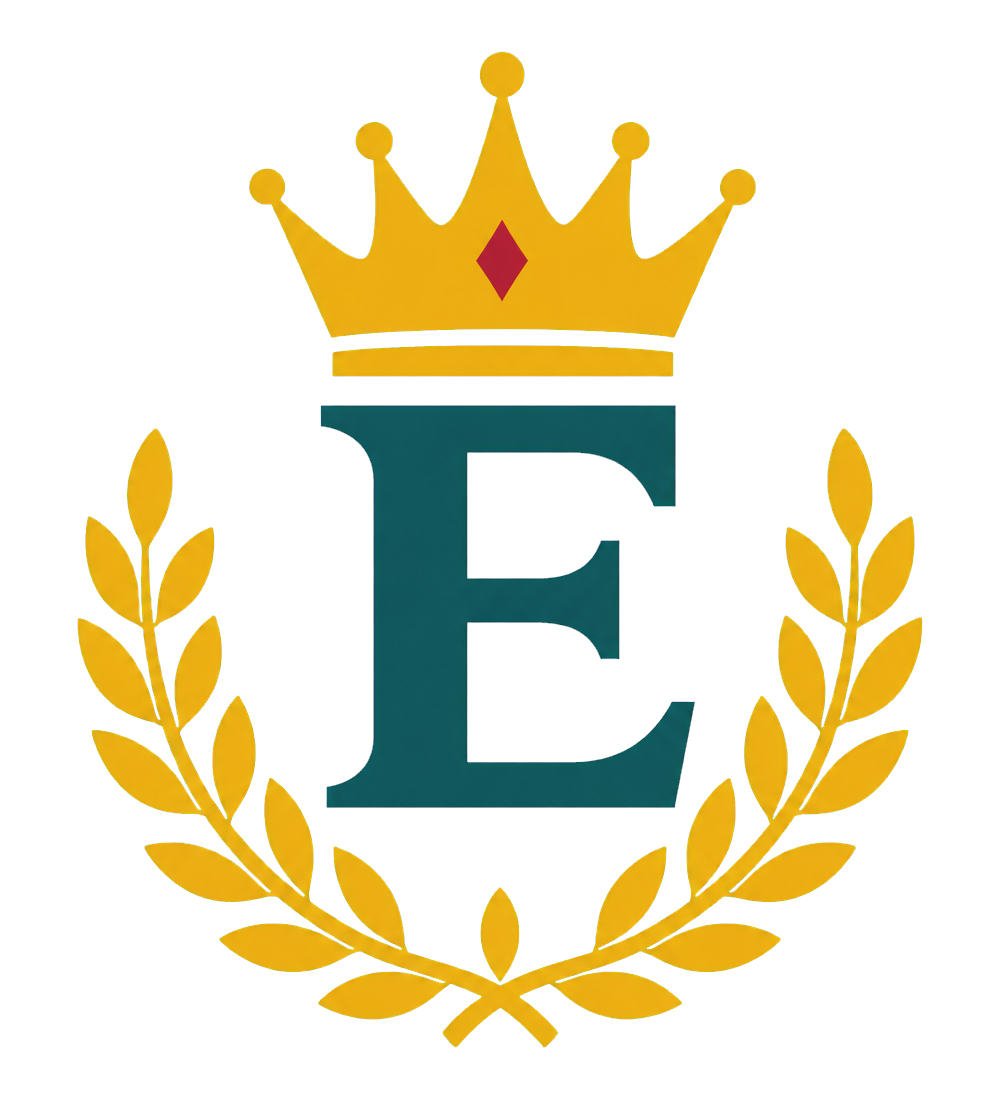}}%
}

\title{\methodlogo\hspace{0.18em}MPIRE: Explicit Manipulation Planning as a Learnable Intermediate Representation for Egocentric Hand-Motion Forecasting}

\author{
    Wen Wang\textsuperscript{1,2},
    Ruibing Hou\textsuperscript{1}\corresponding,
    Hong Chang\textsuperscript{1,2},
    Shiguang Shan\textsuperscript{1,2},
    Xilin Chen\textsuperscript{1,2}
}
\affiliations{
    \textsuperscript{1}Key Laboratory of Intelligent Information Processing of Chinese Academy of Sciences (CAS), Institute of Computing Technology, CAS, China\\
    \textsuperscript{2}University of Chinese Academy of Sciences, China\\
    Correspondence to: Ruibing Hou \texttt{houruibing@ict.ac.cn}
}

\begin{document}

\maketitle

\input{sections/abstract}


\input{sections/introduction}
\input{sections/related_work}
\input{sections/method}
\input{sections/data}
\input{sections/experiments}
\input{sections/conclusion}
\bibliography{paper_refs}

\newpage
\appendix
\input{sections/appendix}

\end{document}

%% file: sections/abstract.tex
\begin{abstract}
Forecasting dexterous hand motions from egocentric observations is fundamental to intelligent interactive systems. Existing VLM-based methods typically map observations directly to future motions, overlooking the underlying manipulation process that governs hand-object interactions. Moreover, end-to-end optimization couples manipulation learning with motion synthesis, causing motion-generation gradients to interfere with the pre-learned manipulation-aware representations.
To overcome these limitations, we propose \methodname{}, a two-stage framework that introduces \textbf{E}xplicit \textbf{M}anipulation \textbf{P}lanning as an \textbf{I}ntermediate \textbf{R}epresentation for \textbf{E}gocentric hand-motion forecasting. \textbf{\textit{Stage~I: Learn to Plan}}. \methodname{} first learns explicit manipulation plans from multimodal context to capture the hand-object interactions progression.
\textbf{\textit{Stage~II: Learn to Act}}. A motion generator synthesizes future bimanual-hand motions conditioned on frozen plan's representations, preventing motion-generation gradients from affecting manipulation planning
To support our method, we further construct \textbf{EMPIRE-651K}, a bimanual hand-motion forecasting dataset comprising $650{,}910$ training windows across $111$ tasks, each paired with an explicit per-hand manipulation plan.
Under identical training and evaluation protocols, \methodname{} achieves state-of-the-art forecasting accuracy,, with an MPJPE of $84.53$~mm and a finger-relative error of $38.97$~mm.
We release the code and Dataset at \url{https://github.com/wangwen-banban/EMPIRE}.
\end{abstract}

%% file: sections/introduction.tex
\section{Introduction}
Forecasting future dexterous hand movements from egocentric observations is fundamental to intelligent interactive systems. It has broad applications in robot learning, virtual and augmented reality, and human--robot collaboration. With the growing availability of large-scale human videos, recent studies have explored their potential as scalable sources of semantic and physical knowledge for embodied manipulation learning \citep{feng2026humanvideos}. Unlike holistic human-motion prediction, egocentric hand-motion forecasting focuses on fine-grained bi-manual coordination and complex finger articulation during object manipulation. Such intricate dynamics make accurate long-horizon forecasting particularly challenging. Nevertheless, recent advances in large-scale egocentric datasets and vision--language foundation models have driven significant progress in this field \citep{ego4d,egodex,paligemma2,qwen3vl}.

\begin{figure}[t]
\centering
\includegraphics[width=0.92\columnwidth]{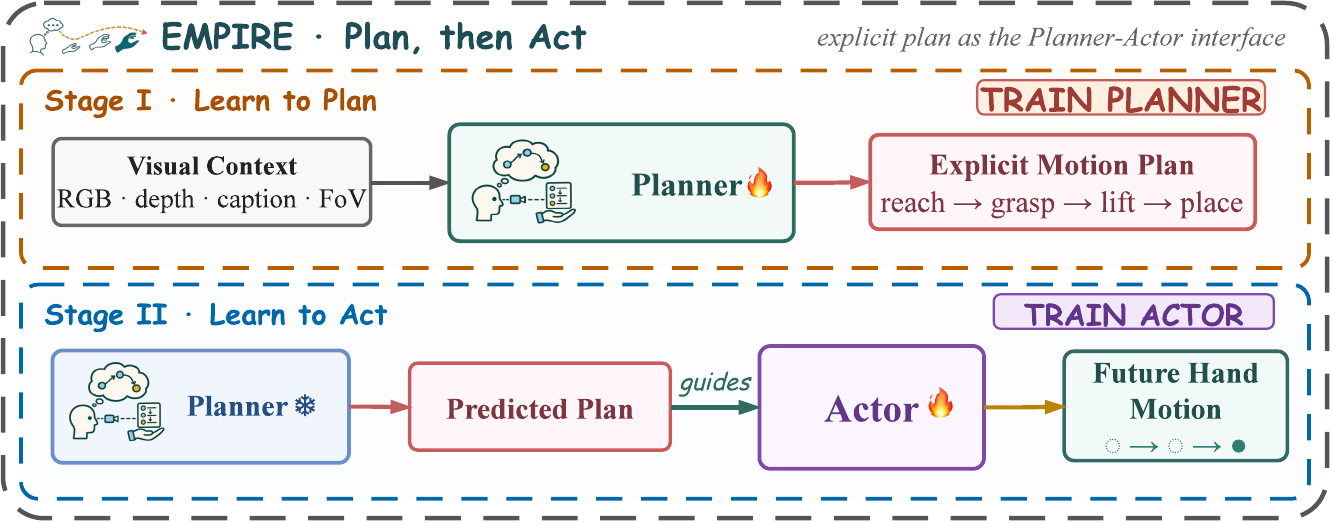}
\caption{\textbf{\methodname{} at a glance.} Stage~I learns to plan; Stage~II learns to act with the planner frozen.}
\label{fig:teaser}
\end{figure}

Hand--object motion generation has consequently shifted from text-conditioned synthesis \citep{goal,imos,text2hoi,diffh2o,hoigpt} towards vision--language-guided forecasting \citep{vitra,beingh0,megohand}, where egocentric observations provide the scene, object, and viewpoint cues which are unavaiable from  language alone. These approaches mainly follow two paradigms. Diffusion-based methods such as VITRA \nocite{vitra} generate continuous hand motion conditioned on a VLM representation, whereas autoregressive models such as Being-H0 \nocite{beingh0} jointly model multimodal observations and discretized future hand motion in a unified sequence. Despite their architectural differences, these methods share a common design: directly mapping VLM representations to future hand motion without explicitly modeling the intermediate manipulation process that drives hand-motion generation.

While recent progress, existing methods remain fundamentally constrained by two overlooked design limitations. \textbf{1) Implicit supervision of manipulation planning.} Future hand movements are generated through a sequence of manipulation decisions, including object interaction, hand coordination, and task progression. Accurate forecasting therefore requires understanding not only how the hand moves but also what manipulation process should occur next. Existing methods, however, supervise the final trajectory or motion-token sequence, leaving manipulation planning to be inferred implicitly during motion prediction. Without an explicit representation of the manipulation process, generated motion may remain locally plausible while deviating from the intended interaction progression, particularly in long-horizon tasks where early planning errors accumulate over time.

\textbf{2) Coupled optimization of semantic learning and motion generation.} 
Existing egocentric hand forecasting methods typically jointly optimize the VLM backbone and the motion generation model \citep{rt2,openvla,pi0}. Consequently, motion-generation gradients continually update the semantic representation used for conditioning, 
causing the generator to adapt to a dynamically changing feature space.
Meanwhile, optimizing directly for spatial motion objectives may overwrite manipulation-aware structure acquired during VLM pretraining, weakening the representation required for scene understanding and interaction anticipation.  Therefore, this coupled optimization may limit the ability of VLMs to provide stable semantic guidance and ultimately constrain forecasting performance.

As summarized in Figure~\ref{fig:teaser}, we introduce \textbf{\methodname{}}, a two-stage method that first learns Explicit Manipulation Planning as an Intermediate Representation and then leverages it to forecast future hand motion from Egocentric-view. In \textbf{\textit{Stage~I: Learn to Plan}}, a VLM predicts an explicit, motion-oriented manipulation plan from an egocentric RGB observation, RGB-derived monocular depth, 
a coarse instruction, and camera field of view. Monocular depth provides complementary geometric context because general-purpose VLMs remain unreliable at metric distance and complex 3D spatial reasoning \citep{spatialvlm}. 
Rather than 
predicting hand motion, the plan decomposes the anticipated interaction into temporally ordered steps for both hands, providing an interpretable intermediate representation. In \textbf{\textit{Stage~II: Learn to Act}}, the entire planner is frozen, and a flow-matching DiT \citep{lipman2023flow} 
predicts future two-hand motion from the current hand state and the hidden representation of the predicted plan. Training on plans predicted by Stage~I rather than ground-truth plans matches the condition available at inference and reduces the train--deployment gap. By preventing motion gradients from updating the planner, this decoupled design preserves manipulation-aware representations, provides stable semantic conditioning.

For training, we construct \textbf{EMPIRE-651K} from EgoDex~\citep{egodex} by augmenting $650{,}910$ windows across $111$ manipulation tasks with explicit per-hand manipulation plans. Extensive experiments demonstrate that \methodname{} achieves accurate and efficient egocentric hand-motion forecasting, improving both global hand placement and fine-grained finger articulation. It consistently outperforms existing methods under a unified evaluation protocol, reducing MPJPE by \textbf{19.8\%} compared with VITRA while decreasing optimization time by \textbf{38.8\%}. Moreover, \methodname{} achieves comparable accuracy to the much larger Being-H0-14B model with \textbf{83.5$\times$ faster} end-to-end inference. 




%% file: sections/related_work.tex
\section{Related Work}

\subsection{Vision--Language Models}
Vision--language models (VLMs), enabled by visual instruction tuning and large-scale multimodal pretraining, have demonstrated strong open-vocabulary scene understanding from visual and linguistic inputs \citep{llava,paligemma2,internvl,qwen25vl,qwen3vl}. However, general-purpose VLMs remain limited in metric-scale perception and fine-grained 3D spatial reasoning, which are critical for spatial interaction \citep{spatialvlm,vsibench,spatial457}. To bridge this gap, recent embodied systems leverage VLMs as high-level planners. For example, SayCan grounds language-model plans with environmental affordances \citep{saycan}, PaLM-E enables multimodal sequential planning for embodied tasks \citep{palme}, RT-H predicts language-conditioned motion intentions before execution \citep{rth}, and Embodied Chain-of-Thought introduces intermediate reasoning over plans, subtasks, motions, and visual grounding \citep{ecot}. These studies demonstrate the potential of VLMs for embodied planning. 

\subsection{Hand Motion Generation and Forecasting}
Hand Motion Generators connect multimodal perception to hand actions. RT-2 and OpenVLA predict robot actions from visual-language inputs \citep{rt2,openvla}, Octo learns a generalist policy across heterogeneous embodiments \citep{octo}, and $\pi_0$ integrates a pretrained VLM with a flow-based action expert for action generation \citep{pi0}. XL-VLA further extends this paradigm by adopting a $\pi_0$-style flow-based action expert in a shared latent action space across heterogeneous dexterous robotic hands \citep{jiang2026crosshand}. 
For dexterous hand-motion generation, VITRA conditions a DiT-based motion generator on learned VLM representations \citep{vitra}, Being-H0 autoregressively predicts discretized motion tokens conditioned on multimodal observations \citep{beingh0}, and MEgoHand combines VLM-derived motion priors, monocular depth, and a flow-matching policy for egocentric hand control \citep{megohand}. 
These advances are enabled by increasingly large-scale interaction datasets, including GRAB with whole-body grasping and object meshes \citep{grab}, H2O and ARCTIC with hand--object manipulation sequences \citep{h2o,arctic}, HOT3D with multi-view egocentric 3D hand-object tracking \citep{hot3d}, and EgoDex with large-scale egocentric manipulation videos and tracked 3D hands \citep{egodex}.
Despite these advances, existing hand-motion forecasting approaches typically generate future motions directly from implicit multimodal representations, without explicitly modeling the intermediate manipulation process that governs hand-object interactions.

%% file: sections/method.tex
\section{Method}
\providecommand{\planablationref}{the supplementary analysis of Stage~II training with predicted versus ground-truth plans}

\begin{figure*}[t]
\centering
    \includegraphics[width=\textwidth]{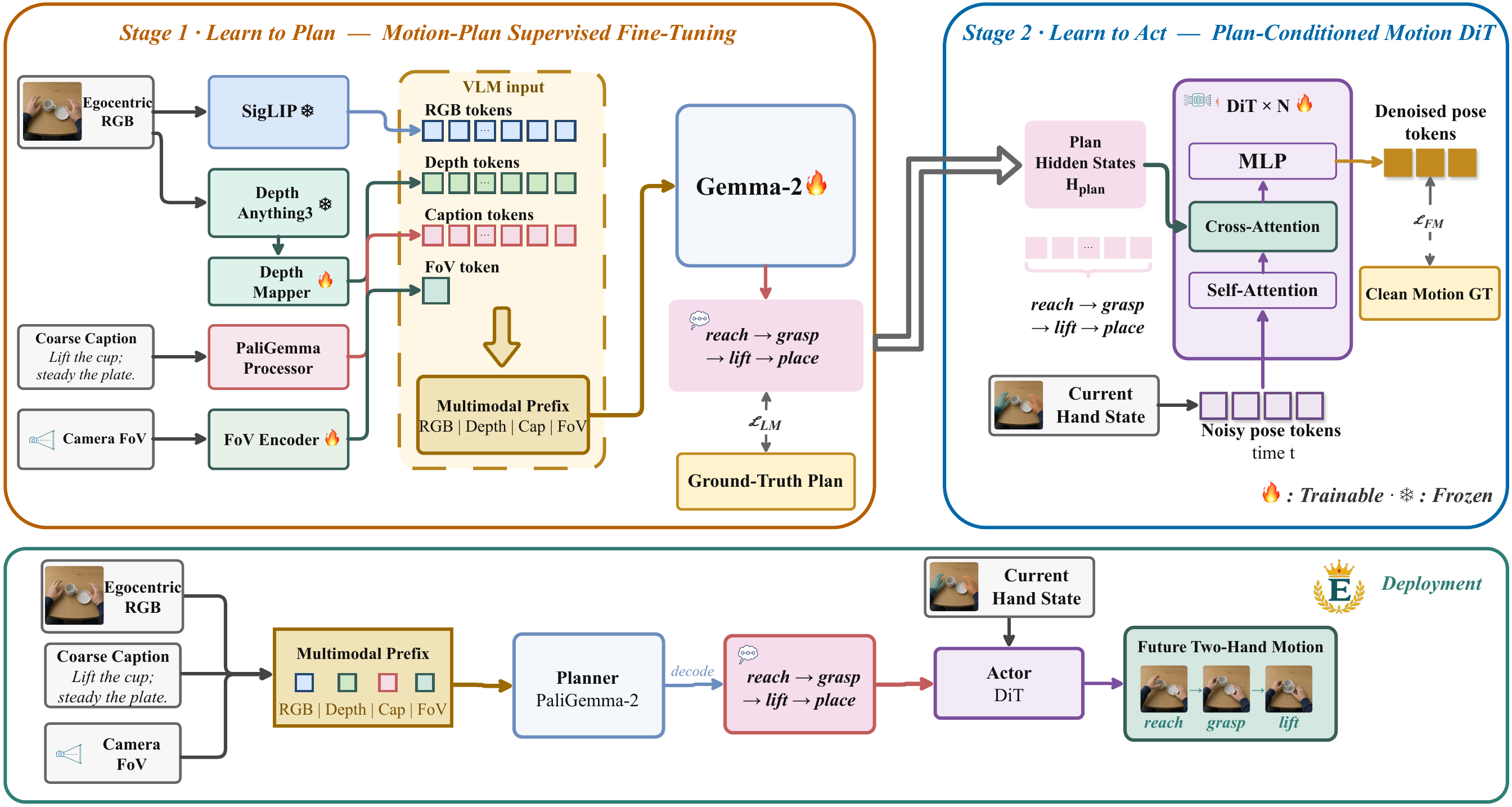}
\caption{\textbf{Overview of \methodname{}.}
\emph{Stage I (left):} RGB and inferred monocular-depth features, together with the caption and camera FoV, form a multimodal prefix that conditions explicit motion-plan generation.
\emph{Stage II (right):} The entire Stage-I VLM is frozen, while the 
motion generator is trained with flow matching by conditioning on the current hand state and cross-attending to the hidden states $H_{\mathrm{plan}}$ of Stage-I prediction.
\emph{Bottom:} During inference, the frozen VLM generates a plan with hidden states, which directly condition the DiT to generate future motion.}
\label{fig:method}
\end{figure*}

As shown in Figure~\ref{fig:method}, \methodname{} follows a two-stage plan-then-act pipeline composed of a Planner and an Actor. The Planner generates an explicit manipulation plan from multimodal egocentric observations, and the Actor transforms the plan hidden states into future two-hand motion.

\subsection{Problem Formulation}
We formulate egocentric dexterous hand-motion forecasting as predicting future bimanual motion from multimodal egocentric observations. Let $O=(I,c,f)$ denote an egocentric RGB observation $I$, a coarse action caption $c$, and the camera field of view $f$. Given $O$ and the current two-hand state $s_0$, the goal is to generate a future trajectory $A=(a^{(1)},\ldots,a^{(T)})\in\mathbb{R}^{T\times d_a}$, where $T$ is the prediction horizon and $d_a$ is the dimension of the two-hand pose at each step. The generated trajectory should remain continuous with the current state $s_0$ while consistent with the manipulation context described by $O$.

Directly learning $p(A\mid O,s_0)$ leaves the manipulation process that connects perception to motion implicitly. We instead introduce a structured motion plan $P$ as an intermediate representation. The plan describes the anticipated interaction as temporally ordered, hand-specific sub-actions, while $H_P$ denotes the hidden states of the Planner's tokens after the Planner has conditioned on $O$. These hidden states serve as the interface between planning and motion generation, yielding the following two-stage decomposition:
\begin{equation}
p(A\mid O,s_0)
=
\sum_P
p_{\theta}(A\mid H_P,s_0)\,
p_{\phi}(P\mid O),
\label{eq:plan-act-factorization}
\end{equation}
where $p_{\phi}(P\mid O)$ denotes the Planner learned in Stage~I and $p_{\theta}(A\mid H_P,s_0)$ denotes the Actor learned in Stage~II. During deployment, the latent plan variable is instantiated by the Planner prediction, 
whose hidden states are subsequently consumed by the Actor.

\subsection{Stage I: Learn to Plan}
Stage~I instantiates the Planner $p_{\phi}(P\mid O)$ with a PaliGemma-2 VLM \citep{paligemma2}.
Given the observation $O$, we construct a 
multimodal prefix from RGB appearance \citep{siglip}, inferred monocular depth feature \citep{depthanything3}, caption, and horizontal and vertical FoV:
\begin{equation}
Z=[Z_{\mathrm{rgb}};Z_{\mathrm{depth}};Z_{\mathrm{cap}};Z_{\mathrm{fov}}],
\label{eq:multimodal-prefix}
\end{equation}
in a fixed order. RGB tokens provide semantic appearance cues, while depth tokens expose complementary scene geometry. Caption and FoV tokens specify the task and camera configuration, respectively.

The supervision target is a motion-oriented plan $P=(P_1,\ldots,P_{|P|})$ which decomposes the coarse action caption into temporally ordered, hand-specific sub-actions, such as reaching, stabilizing, grasping, lifting, and placing. 
We mask the multimodal prefix from the language loss and optimize autoregressive next-token prediction only over the plan: 
\begin{equation}
\mathcal{L}_{\mathrm{plan}}
=-\sum_{i=1}^{|P|}\log p_{\phi}(P_i\mid Z,P_{<i}).
\label{eq:plan-objective}
\end{equation}
Here, $p_{\phi}$ denotes the Planner's next-token distribution.
During training, the visual and depth encoders remain frozen, while the planner and lightweight modality adapters are optimized. Therefore, Stage I learns both an explicit manipulation plan and its corresponding hidden representation, which serves as the planning interface for the Actor.

\subsection{Stage II: Learn to Act}
Stage~II instantiates the Actor $p_{\theta}(A\mid H_P,s_0)$ with a flow-matching diffusion transformer. Before Actor training, the Stage-I Planner generates a predicted plan $\hat P$ for each training window which is cached offline as the planning condition. The ground-truth plan $P$ is only used for Stage-I supervision. 
Conditioning Stage II on planner-generated plans rather than ground-truth plans eliminates the mismatch between training-time and deployment-time conditions (see \planablationref).

We initialize the Planner with the Stage-I learned parameters and freeze it during Stage-II training. For each Stage-II example, the frozen Planner processes the multimodal prefix together with the cached predicted plan:
\begin{equation}
H=[h_i]_{i=1}^{|Z|+|\hat P|}
=\mathcal{M}_{\phi}(Z,\hat P),
\end{equation}
where $\mathcal{M}_{\phi}$ denotes the VLM planner. 
We extract the hidden states corresponding to the generated plan span:
\begin{equation}
H_{\mathrm{plan}}=[h_i]_{i=|Z|}^{|Z|+|\hat P|}
\end{equation}
as the manipulation representation. A lightweight projector $\mathcal{G}_{\psi}$ maps this variable-length span to the Actor conditioning space, while $\mathcal{E}_{\mathrm{state}}$ encodes the current hand state:
\begin{equation}
Z_{\mathrm{state}}=\mathcal{E}_{\mathrm{state}}(s_0),
\qquad
Z_{\mathrm{plan}}=\mathcal{G}_{\psi}(H_{\mathrm{plan}}).
\end{equation}
The DiT action expert treats noisy motion tokens as queries and cross-attends to $Z_{\mathrm{plan}}$ as keys and values; $Z_{\mathrm{state}}$ supplies the initial-pose condition. Thus, the Actor generates future motion conditioned on the Planner’s manipulation-aware representation rather than directly relying on an entangled image-caption feature.

We train the DiT-based Actor \citep{peebles2023dit} and its condition projector with flow matching \citep{lipman2023flow}. Given a ground-truth trajectory $A^\star$, Gaussian noise $\epsilon\sim\mathcal{N}(0,\mathbf{I})$, and $\tau\sim\mathcal{U}(0,1)$, we construct the linear interpolation
\begin{equation}
A_{\tau}=(1-\tau)\epsilon+\tau A^\star
\end{equation}
and optimize the Actor to predict the velocity field:
\begin{equation}
\mathcal{L}_{\mathrm{act}}
=\mathrm{E}_{\epsilon,A^\star,\tau}\left[
\left\|\mathcal{V}_{\theta}(A_{\tau},\tau,Z_{\mathrm{state}},Z_{\mathrm{plan}})
-(A^\star-\epsilon)\right\|_2^2
\right],
\label{eq:actor-objective}
\end{equation}
where $\mathcal{V}_{\theta}$ denotes the Actor. 
The Actor predicts the velocity field over the full two-hand trajectory instead of separately modeling the two hands. Freezing the Planner prevents the motion generation objective from altering the learned planning representation, while avoiding back-propagation through the Planner.

\subsection{Deployment: Plan Then Act}
At deployment, the frozen Stage-I Planner first predicts the structured motion plan $\hat P$ online. We prefill the VLM with $Z$ and retain the final-layer hidden state of each 
generated plan token, where the predefined plan delimiters identify the span used for extracting $H_{\mathrm{plan}}$. The Stage-II Actor then conditions on this cached span together with $s_0$. The explicit plan is therefore generated once, and provides a stable, manipulation-aware representation for motion generation.


%% file: sections/data.tex
\raggedbottom
\begin{figure*}[h]
\centering
\includegraphics[width=\textwidth]{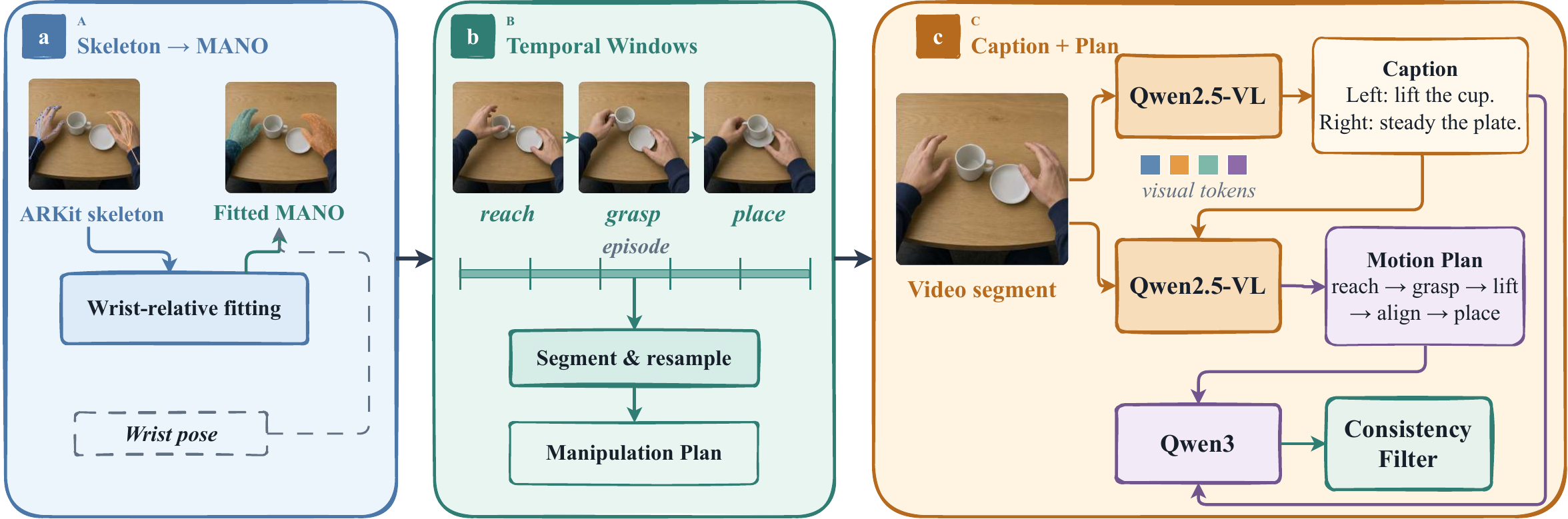}
\caption{\textbf{The EMPIRE-651K construction and plan-supervision pipeline.} (A) Raw ARKit skeletons are converted into wrist-aligned MANO motion targets. (B) Multi-task EgoDex videos are split into task-specific episodes, which are then segmented and resampled into aligned forecasting windows. (C) Qwen2.5-VL first captions each video segment and then generates a caption-grounded motion plan from the segment and caption; Qwen3 labels caption--plan consistency, and inconsistent samples are discarded.}
\label{fig:data-pipeline}
\end{figure*}

\section{Dataset: EMPIRE-651K}
\label{sec:dataset}
To enable long-horizon egocentric bimanual motion forecasting with explicit manipulation plans, we construct \textbf{EMPIRE-651K} from the EgoDex dataset \citep{egodex}. EgoDex contains  $829$ hours of egocentric Apple Vision Pro recordings at $30$~FPS across $194$ tabletop manipulation tasks, with synchronized RGB observations, camera calibration, language task descriptions, and two-hand ARKit skeleton annotations. Following the official split, we use the five training partitions for dataset construction and keep the test partition completely held out.
After processing below, EMPIRE-651K contains $650{,}910$ five-second forecasting windows across $111$ manipulation tasks. The held-out test partition contains an additional $6{,}836$ windows covering the same task vocabulary. Each sample consists of a current egocentric RGB observation, the initial two-hand state, a task caption, an explicit manipulation plan, and the future two-hand trajectory. Figure~\ref{fig:data-pipeline} shows the dataset construction.

\paragraph{(A) Skeleton-to-MANO conversion.}
 EgoDex provides $21$ tracked 3D hand joints but does not include a parametric hand representation required for motion forecasting. We therefore convert the original skeleton annotations into MANO-based hand representations. Specifically, we transform finger joints into the wrist coordinate frame, resolve left-hand chirality inconsistencies, and perform sequence-level MANO fitting \citep{mano}. The fitting optimizes a $15$-dimensional PCA pose representation with a neutral mean shape $\beta{=}0$, guided by joint reconstruction and first-order temporal smoothness constraints. We set the weights for the reconstruction and smoothness terms to $100$ and $0,2$, 
 respectively, and perform optimization using Adam \citep{kingma2015adam} for 500 iterations with a learning rate of 0.01. The resulting MANO finger articulations are combined with the original ARKit wrist $SE(3)$ transformations, preserving both global hand motion and fine-grained finger articulation to obtain temporally consistent two-hand motion targets.
 
\paragraph{(B) Temporal window construction.}
EgoDex recordings may contain multiple manipulation episodes within a single video. Directly treating an entire recording as one sequence would introduce task transitions unrelated to the target action, and weaken the correspondence between observation, instruction, and future motion. We therefore first divide each recording into task-specific episodes and further partition each episode into non-overlapping five-second windows.
Each window contains 150 frames at the original 30 FPS and is resampled to 12 FPS for motion forecasting. Under our forecasting setting, the model receives only the current egocentric observation and initial two-hand state, while predicting the following 60 frames of bimanual motion. Each example therefore requires forecasting a full five seconds of future motion, enabling the study of long-horizon manipulation evolution rather than short-term motion continuation.

\paragraph{(C) Caption and motion-plan annotation.}
To supervise explicit manipulation planning, we construct task captions and manipulation plans through a two-stage VLM-assisted annotation pipeline. Given each five-second forecasting segment, the pipeline first identifies the overall manipulation intent and then decomposes it into temporally ordered, hand-specific sub-actions.
\textbf{i) Caption generation.} 
For each five-second segment, Qwen2.5-VL-7B-Instruct \citep{qwen25vl} observes video frames sampled at $3$~FPS and generates a concise caption describing the dominant manipulation intent and involved interactions. The caption provides coarse semantic guidance without requiring detailed descriptions of hand motion evolution.
\textbf{ii) Manipulation plan generation.} 
Conditioned on the video segment and generated caption, Qwen2.5-VL produces a structured manipulation plan that decomposes the interaction into temporally ordered, hand-specific sub-actions, such as reaching, grasping, stabilizing, lifting, and placing. The caption acts as a semantic constraint to maintain consistency with the observed task and reduce unsupported actions.
\textbf{iii) Annotation filtering.} 
To improve annotation reliability, Qwen3-30B-A3B-Instruct-2507 \citep{qwen3} evaluates the consistency between each caption and its corresponding manipulation plan using binary labels. Samples with inconsistent annotations are removed. The detailed prompts for caption generation, plan construction, and consistency evaluation are provided in the supplementary material under \emph{Data-Annotation Prompts}.

\paragraph{Human annotation audit.}
To evaluate annotation quality, we conduct a human audit over randomly sampled training windows. We select 20 samples from each EgoDex training partition, resulting in 100 manually inspected cases from distinct manipulation episodes.
Each sample is evaluated according to six criteria, including caption grounding, hand attribution, plan grounding, plan coverage, temporal coordination, and usefulness for motion forecasting. All sampled videos are successfully interpretable. As shown in Table~\ref{tab:human-audit}, the final annotation quality reaches 90.41/100, and 84.0\% of caption-plan pairs are considered suitable for downstream learning.  These results indicate that the generated captions and manipulation plans are sufficiently grounded in the observed interactions and provide reliable supervision for learning the planning interface between perception and motion generation.
Detailed evaluation protocols and criterion definitions are provided in the supplementary material 

\begin{table}
\centering
\small
\setlength{\tabcolsep}{5pt}
\begin{tabular}{lc}
\toprule
Metric & Score \\
\midrule
Caption quality & $4.57$ \\ 
Motion-plan quality & $4.66$ \\
Overall quality & $\mathbf{4.62}$ \\
Usable pairs & $\mathbf{84.0\%}$ \\
\bottomrule
\end{tabular}
\caption{\textbf{Human audit of EMPIRE-651K annotations.} Final results over $100$ stratified training windows. Quality scores use a $1$--$5$ scale; A pair is usable when both its caption and plan means are at least $4/5$.}
\label{tab:human-audit}
\end{table}

%% file: sections/experiments.tex
\begin{table*}[t]
\centering
\small
\setlength{\tabcolsep}{10pt}
\begin{tabular}{l c ccc c}
\toprule
Method & Params & MPJPE $\downarrow$ & Wrist MPJPE $\downarrow$ & Finger-
relative MPJPE\ $\downarrow$ & Infer.(s/window)\ $\downarrow$ \\
\midrule
VITRA re-impl. \citep{vitra} & $3.2$B & $105.37$ & $91.42$ & $\underline{45.37}$ & $\mathbf{0.4}$ \\
Being-H0-1B \citep{beingh0} & $1.2$B & $115.61$ & $100.23$ & $66.87$ & $15.6$ \\
Being-H0-8B \citep{beingh0} & $8.2$B & $85.39$ & $\underline{68.61}$ & $50.95$ & $28.3$ \\
Being-H0-14B \citep{beingh0} & $15.4$B & $\underline{84.90}$ & $\mathbf{66.54}$ & $51.65$ & $83.5$ \\
\midrule
\methodname{} (ours) & $3.2$B & $\mathbf{84.53}$ & $73.25$ & $\mathbf{38.97}$ & $\underline{1.0}$ \\
\bottomrule
\end{tabular}
\caption{\textbf{Main results on EMPIRE-651K.} EMPIRE is compared with baselines on the same held-out test set. See the Evaluation subsection for protocol details and metric definitions.}
\label{tab:fulldata}
\end{table*}
\section{Experiment}
\label{sec:experiment}

\subsection{Training}
\paragraph{Data splits.}
The final EMPIRE model and our re-implemented VITRA baseline are trained on all five training partitions of EMPIRE-651K, comprising $650{,}910$ forecasting windows from $111$ manipulation tasks. The held-out test partition is excluded from all training runs.

\paragraph{Implementation Details.}
Our backbone is PaliGemma-2-3B with a $182$M DiT-Base flow-matching generator. The model is trained following the two-stage framework. In Stage~I, the planner model is fine-tuned for $1$ epoch. In Stage~II, the VLM is frozen and only the DiT action generator is optimized for $4$ epochs. Training  is performed with a total batch size of $64$ across 8 A40 GPUs, more details in the Supplementary material.


\paragraph{Evaluation Protocol.}
All models are evaluated on the held-out test partition containing 6,836 five-second forecasting windows from 111 manipulation tasks. The same test set is used for models trained on all five partitions and for all Part~1 ablations.
For Part~1 ablations, test windows are further categorized as \emph{seen} or \emph{unseen} based on task overlap with the training set. Specifically, 1,247 windows from the 26 training tasks are labeled as seen, while 5,589 windows from 85 unseen tasks are labeled as unseen. This split is only used for Part~1 ablations due to their limited training task coverage.
And since the motion generator is stochastic, we sample $K{=}8$ trajectories per input and report best-of-$K$ performance following prior motion-generation works \citep{vitra,beingh0}. Each trajectory is generated with $4$ Euler steps, using the same sampling protocol for all models.

\paragraph{Metrics.}
All methods are evaluated using MPJPE and its variants. Errors are reported in millimeters, with lower values indicating better performance. Unless otherwise specified, global-coordinate metrics are computed in the absolute camera coordinate frame. \textbf{i) MPJPE} measures the average Euclidean distance over both hands, all valid prediction frames, and all MANO joints. Under the best-of-8 setting, we select the trajectory with the lowest MPJPE for each test sample and compute all other metrics based on the selected trajectory.  \textbf{ii) Wrist MPJPE} measures global hand placement accuracy by averaging the error of wrist joints over time. \textbf{iii) Finger-relative MPJPE} removes wrist translation before computing joint errors to evaluate finger articulation independent of global hand motion.


\subsection{Main Results}

\begin{figure}[h]
\centering
\includegraphics[width=\columnwidth]{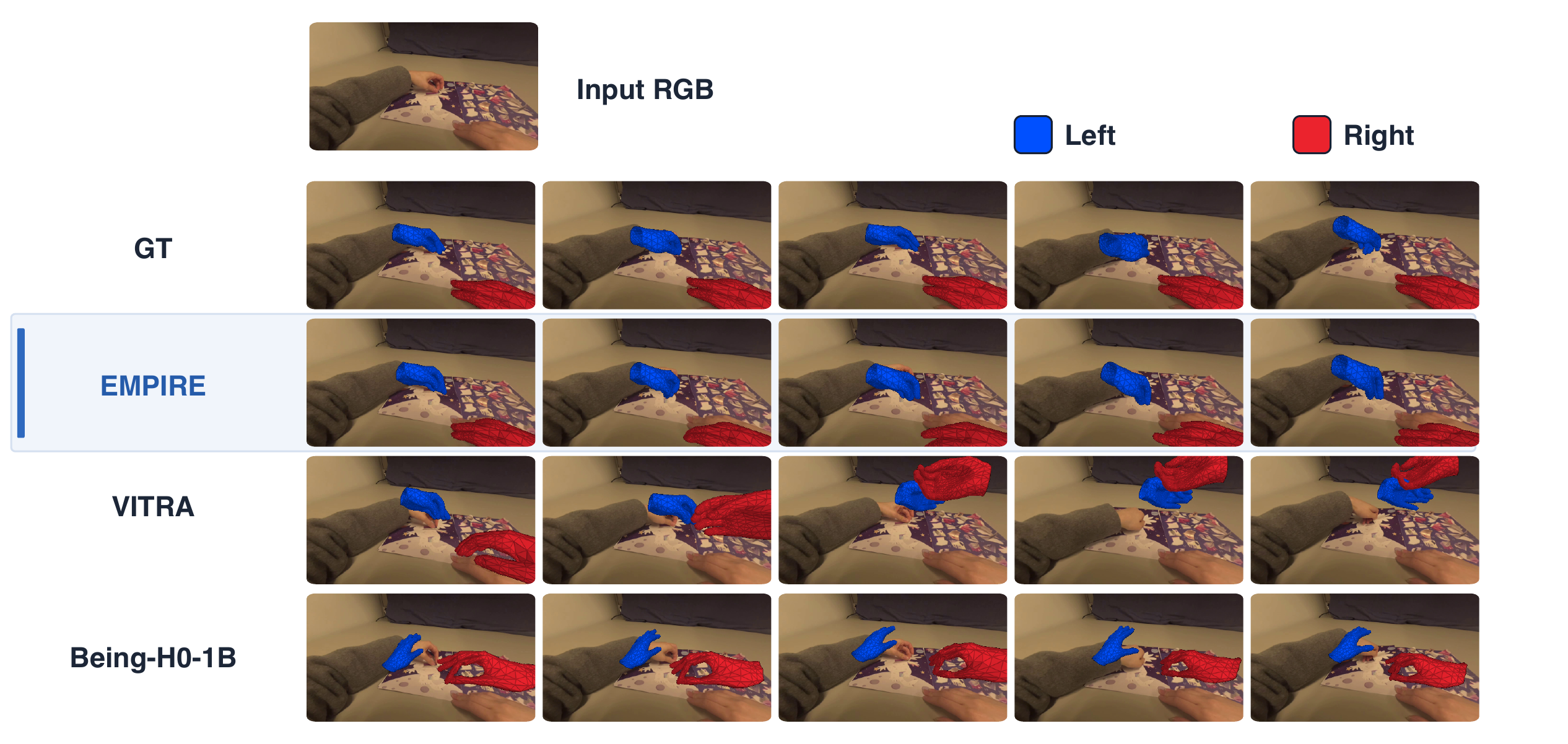}
\caption{Qualitative comparison of motion predictions against baselines. \texttt{Task: peel\_place\_sticker}.}
\label{fig:qual-motion-case1}
\end{figure}

\paragraph{Forecasting accuracy.}
Table~\ref{tab:fulldata} compares EMPIRE with VITRA  \cite{vitra} and Being-H0 \cite{beingh0}. EMPIRE achieves the best overall performance among the evaluated methods, reducing MPJPE by 19.8\% compared with VITRA while using a substantially smaller model than the larger Being-H0 variants. Although the largest Being-H0 model achieves better wrist localization, EMPIRE obtains substantially more accurate finger articulation, demonstrating the advantage of explicit manipulation planning for fine-grained hand motion forecasting over long horizons. Figure~\ref{fig:qual-motion-case1} shows the same trend qualitatively: EMPIRE remains stable over the long horizon, while the other methods are relatively stable early but degrade at later steps. Additional cases are provided in the supplementary material under \emph{Qualitative Hand-Motion Forecasts}.

\paragraph{Training and inference efficiency.}
Beyond forecasting accuracy, EMPIRE also achieves improved efficiency. Training the complete two-stage framework requires 71 hours on 8 A40 GPUs, compared with 116 hours for the VITRA re-implementation, reducing training time by 38.8\%. During inference, EMPIRE generates each forecasting window in 1.0 second, substantially faster than Being-H0 models. In particular, EMPIRE is $83.5\times$ faster than Being-H0-14B in inference, while achieving comparable overall MPJPE and substantially better finger articulation accuracy. Detailed latency analysis is provided in the supplementary material under \emph{Inference Cost}.

\paragraph{When does motion planning help?}
To investigate when explicit planning is most beneficial, we analyze planning gains across different task difficulties. Table~\ref{tab:plan-difficulty} groups test tasks according to their baseline MPJPE and shows that the benefit of planning increases with task complexity. The average MPJPE reduction grows from 4.6 mm on easy tasks to 49.7 mm on hard tasks, indicating that explicit plans are particularly effective for challenging manipulation scenarios. 
Overall, planning improves 89 out of 111 tasks, achieving an average MPJPE reduction of 22.3 mm. 
This trend is further supported by the negative correlation between baseline MPJPE and planning gain ($r=-0.567$) \citep{pearson1895}, suggesting that tasks with larger initial errors benefit more from explicit planning.
Furthermore, planning generalizes beyond the training task categories, improving 74 out of 85 unseen tasks and 15 out of 26 seen tasks. These results suggest that explicit manipulation plans provide structured temporal guidance, which is particularly valuable for long-horizon and compositional manipulation tasks.

\begin{table}[t]
\centering
\begin{tabular}{@{}lccc@{}}
\toprule
\shortstack[l]{Difficulty\\(baseline MPJPE)} & \#tasks & Mean $\Delta$ & Plan helps \\
\midrule
Easy ($<100$) & $43$ & $-4.6$ & $25/43$ \\
Medium ($100$--$175$) & $49$ & $-26.2$ & $47/49$ \\
Hard ($\geq175$) & $19$ & $-49.7$ & $17/19$ \\
\midrule
All & $111$ & $-22.3$ & $89/111$ \\
\bottomrule
\end{tabular}
\caption{\textbf{Planning benefit by task difficulty.} $\Delta$ means that ours minus baseline; Negative means improvement.}
\label{tab:plan-difficulty}
\end{table}

\subsection{Ablation and Discussion}
To ensure computationally efficient and controlled ablation studies, we conduct all ablations on $26$-task Part~1 subset. 
This setting enables evaluation of both in-distribution performance and generalization to the 85 tasks excluded from ablation training. Unless otherwise specified, each ablation modifies only the component indicated by its name, while keeping the input representation, motion targets, optimization schedule, and evaluation protocol fixed.

\paragraph{Model components.}
Table~\ref{tab:main} evaluates the contribution of explicit motion plans and monocular depth features. Starting from the VITRA re-implementation with caption-only conditioning, introducing the explicit motion plan reduces overall MPJPE from 124.52 mm to 102.26 mm. The improvement is particularly large on unseen tasks, where the error decreases from 129.73 mm to 104.80 mm, suggesting that the plan provides transferable temporal guidance beyond the semantic information contained in the caption alone. This result highlights the importance of explicitly modeling the intermediate manipulation process rather than directly mapping observations and instructions to future hand motions.

We further evaluate different strategies for incorporating depth information. Direct depth concatenation achieves the best performance, reducing overall MPJPE to 95.42 mm, while cross-attention over the same depth features does not improve over the plan-only model. This indicates that geometric cues are more effective when directly integrated into the multimodal representation. Interestingly, the depth-augmented model mainly benefits unseen tasks, 
suggesting a trade-off between cross-task generalization and fitting to the smaller ablation training subset.

\begin{table}[t]
\centering
\small
\setlength{\tabcolsep}{4pt}
\begin{tabular}{l ccc}
\toprule
Configuration & Overall $\downarrow$ & Seen $\downarrow$ & Unseen $\downarrow$ \\
\midrule
VITRA re-impl. (caption only) & $124.52$ & $92.71$ & $129.73$ \\
$+$ motion plan & $102.26$ & $\mathbf{90.87}$ & $104.80$ \\
$+$ depth (cross-attention) & $104.71$ & $110.10$ & $103.51$ \\
\textbf{$+$ depth (concatenation)} & $\mathbf{95.42}$ & $99.33$ & $\mathbf{94.55}$ \\
\bottomrule
\end{tabular}
\caption{\textbf{Model-component ablation.} All results in the table are reported using MPJPE.}
\label{tab:main}
\end{table}

\paragraph{Preventing motion gradients from updating the VLM.}
Table~\ref{tab:freeze-vlm} investigates whether isolating motion gradients is necessary beyond the two-stage training strategy. In the two-stage trainable setting, updating the VLM with motion gradients results in an MPJPE of 118.44 mm. In contrast, freezing the VLM after Stage~I and optimizing only the motion generator reduces MPJPE to 94.67 mm, improving by 23.77 mm (20.1\%). This result indicates that freezing is not merely a constraint, but preserves stable semantic representations and provides a consistent interface for motion generation.

\begin{table}[t]
\centering
\small
\setlength{\tabcolsep}{4pt}
\begin{tabular}{l c c c}
\toprule
Setting & VLM grad.? & LM loss & MPJPE $\downarrow$ \\
\midrule
Single-stage & Yes & -- & $124.52$ \\
Two-stage, trainable & Yes & $0.5$ & $118.44$ \\
\textbf{Two-stage, frozen (ours)} & \textbf{No} & $0$ & $\mathbf{94.67}$ \\
\bottomrule
\end{tabular}
\caption{\textbf{Motion-gradient isolation in Stage~II.} The plan loss coefficient is set to 0.5.}
\label{tab:freeze-vlm}
\end{table}

\paragraph{DiT capacity.}
Table~\ref{tab:dit-size} studies the effect of actor capacity while fixing plan-span conditioning, concatenated depth features, and a frozen VLM. DiT-M consistently underperforms across all evaluation settings, indicating insufficient capacity to capture complex bimanual hand motion dynamics. Increasing the capacity to DiT-L improves performance on seen tasks but leads to degraded generalization on unseen tasks, suggesting that excessive model capacity may overfit the observed manipulation patterns under the available training data scale. In contrast, DiT-B achieves the best overall and unseen-task performance, providing a better balance between motion modeling capability and generalization. We therefore select DiT-B (182M parameters) as the default actor.

\begin{table}[t]
\centering
\small
\setlength{\tabcolsep}{6pt}
\begin{tabular}{l c ccc}
\toprule
Action head & Params & Overall $\downarrow$ & Seen $\downarrow$ & Unseen $\downarrow$ \\
\midrule
DiT-M & $46$M & $107.93$ & $114.15$ & $106.55$ \\
DiT-B (default) & $182$M & $\mathbf{95.42}$ & $99.33$ & $\mathbf{94.55}$ \\
DiT-L & $637$M & $100.08$ & $\mathbf{95.37}$ & $101.13$ \\
\bottomrule
\end{tabular}
\caption{\textbf{DiT-capacity ablation.} All results in the table are reported using MPJPE}
\label{tab:dit-size}
\end{table}
\flushbottom

%% file: sections/conclusion.tex
\section{Conclusion}

We presented \methodname{}, a two-stage framework that introduces an explicit motion plan as an intermediate interface between vision-language understanding and long-horizon bimanual hand-motion forecasting. Stage I learns structured, manipulation-aware plans for hand-object interactions, while Stage II trains a flow-matching motion generator conditioned on the learned plans with the VLM frozen. This decoupled optimization prevents gradients from low-level motion generation objectives from disrupting the manipulation-aware representations learned during planning. To support this task, we introduced EMPIRE-651K, a large-scale benchmark constructed from EgoDex by converting skeleton annotations into temporally aligned MANO trajectories, coarse instructions, and motion-plan supervision, covering 650,910 training windows across 111 manipulation tasks. Extensive experiments demonstrate that EMPIRE achieves state-of-the-art forecasting accuracy. These results validate explicit manipulation planning combined with decoupled motion learning as an effective paradigm for accurate and efficient egocentric dexterous-motion forecasting.

%% file: sections/appendix.tex

\section{Data-Annotation Prompts}
\label{app:data-prompts}
We reproduce the three prompts used to construct and filter the caption--plan annotations in EMPIRE-651K. Specifically, Qwen2.5-VL first generates a coarse caption from the five-second egocentric video. The video and generated caption are then jointly provided to the model in a second pass to derive the corresponding motion plan. Finally, Qwen3 evaluates each caption--plan pair to assess semantic consistency and filters out inconsistent annotations.

\paragraph{Caption generation (user prompt).}
\begin{quote}
\small\ttfamily\raggedright\sloppy
Describe hand actions in this video sequence. Focus on what left hand and right hand are doing.
Use very simple sentences. Output format must be exactly:

``Left hand: [action description]. Right hand: [action description or None].''

\medskip
For example:

``Left hand: Pick up the cup. Right hand: None.''

``Left hand: Pour water into the glass. Right hand: Hold the glass.''

``Left hand: None. Right hand: Close the door.''

\medskip
What are the hands doing?
\end{quote}

\paragraph{Caption-grounded motion-plan construction (user prompt).}
\begin{quote}
\small\ttfamily\raggedright\sloppy
You are given an egocentric hand-object manipulation video and a coarse caption.

Caption:
\{caption\}

Generate one fine-grained action decomposition of the hand actions in this clip. The decomposition must be consistent with both the video and the caption.

\medskip
Output exactly one XML-like tag:
\tok{cot}1. [first visible hand-action step] 2. [next hand-action step] 3. [next or final hand-action step]\tok{/cot}

\medskip
Rules:

-- Use a numbered list inside the \tok{cot} tag: 1., 2., 3. and optionally 4. or 5.

-- Each numbered step should describe one clear sub-action in temporal order.

-- Mention left hand and right hand when they are visible or active.

-- Use temporal words such as first, then, while, next, finally when helpful.

-- Describe low-level hand motion, contact, grasp, release, lift, place, open, close, rotate, or stabilize events when visible.

-- If a hand is inactive, missing, or captioned as None, state that briefly.

-- Do not invent objects or actions that contradict the caption.

-- Keep the whole content 40 to 120 words.

-- Do not use markdown bullets, headings, or text outside \tok{cot}.
\end{quote}

\paragraph{Caption--plan consistency labeling (user prompt).}
\begin{quote}
\small\ttfamily\raggedright\sloppy
Determine whether the generated caption and motion plan describe the same hand-object manipulation task.

\medskip
Caption:

\{caption\}

\medskip
Motion plan:

\{motion plan\}

\medskip
Label the pair CONSISTENT when the plan preserves the caption's task, object, and hand roles while providing compatible fine-grained steps. Label it INCONSISTENT when the plan changes the task or object, conflicts with the stated hand roles, or introduces an incompatible goal.

Output exactly one label: CONSISTENT or INCONSISTENT.
\end{quote}
Samples labeled \texttt{INCONSISTENT} are removed before constructing the final aligned supervision tuples.

\section{Human Audit of Annotation Quality}
\label{app:annotation-audit}

\paragraph{Sampling and review.}

The audit evaluates the quality of the training annotations. Using random seed \texttt{20260725}, we randomly sample $20$ windows from each of the five training partitions. The resulting $100$ cases are drawn from $100$ different episodes, preventing the evaluation from being biased toward adjacent windows within the same episode. Each case follows the same temporal configuration as training, consisting of a $60$-frame, $12$~FPS, five-second egocentric video window. A human reviewer jointly examines the source video, caption, and ground-truth motion plan, while predicted plans are withheld during evaluation. All sampled videos contain sufficient visual evidence for reliable assessment.

\begin{table}[t]
\centering
\setlength{\tabcolsep}{2.3pt}
\begin{tabular}{>{\raggedright\arraybackslash}p{0.12\columnwidth} >{\raggedright\arraybackslash}p{0.24\columnwidth} >{\raggedright\arraybackslash}p{0.44\columnwidth} r}
\toprule
Target & Criterion & Definition & Score \\
\midrule
Caption & Grounding & Described actions and objects are consistent with the visual evidence in the video. & $4.38/5$ \\
Caption & Hand attribution & Left/right roles, hand activities, and \texttt{None} assignments are correctly identified. & $4.55/5$ \\
\midrule
Plan & Grounding & Described actions, objects, and contact states are supported by the visual evidence. & $4.69/5$ \\
Plan & Coverage & Key sub-actions and all visibly active hands are covered. & $4.68/5$ \\
Plan & Temporal coordination & Step order, simultaneous motion, and bimanual coordination are correct. & $4.60/5$ \\
Plan & Conditioning utility & The plan is specific, non-redundant, complete, and useful for motion generation. & $4.35/5$ \\
\bottomrule
\end{tabular}
\caption{\textbf{Human-audit criteria and final scores.}} 
\label{tab:audit-criteria}
\end{table}

\paragraph{Criteria and quality measurement.}
Each criterion is rated on a scale from $1$ (highly inconsistent with the video) to $5$ (fully consistent and directly usable). A score of $4$ indicates that the main semantics are correctly captured with only minor issues, whereas a score of $3$ indicates an evident discrepancy requiring correction. Table~\ref{tab:audit-criteria} summarizes the six human-audit criteria and their corresponding average scores. 
Let $C$ denote the average score of the two caption-related criteria and $P$ denote the average score of the four plan-related criteria. We compute the overall audit quality as:
\[
Q_{\mathrm{audit}}
= \frac{C+P}{2}.
\]
An annotation pair is considered usable if both $C\geq4$ and $P\geq4$.
Table~\ref{tab:audit-parts} shows that the overall audit quality reaches $4.62/5$, with a usable-pair rate of $84.0\%$. The quality remains consistently high across all five training partitions, with overall scores ranging from $4.46$ to $4.76$, indicating that the audit results are not dominated by any single subset.

The structured error analysis further shows that the remaining annotation issues are primarily localized to hand attribution and plan details. The most frequent error categories correspond to caption hand/\texttt{None} attribution errors ($9/100$ cases), plan hand-role assignment errors ($8/100$ cases), and hallucinated plan steps ($6/100$ cases). No cases are identified as source-video ambiguous, containing unsupported caption details, or involving coarse, repetitive, or truncated motion plans.

\begin{table}[t]
\centering
\small
\setlength{\tabcolsep}{3.5pt}
\begin{tabular}{lcccc}
\toprule
Partition & Caption & Plan & Overall & Usable ($\%$) \\
\midrule
Part~1 & $4.45$ & $4.51$ & $4.48$ & $75.0\%$ \\
Part~2 & $4.63$ & $4.72$ & $4.68$ & $85.0\%$ \\
Part~3 & $4.80$ & $4.71$ & $4.76$ & $95.0\%$ \\
Part~4 & $4.35$ & $4.56$ & $4.46$ & $75.0\%$ \\
Part~5 & $4.63$ & $4.80$ & $4.71$ & $90.0\%$ \\
\midrule 
Overall & $4.57$ & $4.66$ & $\mathbf{4.62}$ & $\mathbf{84.0\%}$ \\
\bottomrule
\end{tabular}
\caption{\textbf{Human-audit quality scores across training partitions.} Each partition contains 20 cases sampled from distinct episodes. The overall row reports the aggregate results over all 100 audited cases.}
\label{tab:audit-parts}
\end{table}

\begin{figure*}[t]
\centering
\begingroup
\setlength{\fboxsep}{0pt}
\setlength{\fboxrule}{0.35pt}

\small\textbf{(a) Folding a cloth (Part~2; audit case \#11).}\par\vspace{0.25em}
\fbox{\includegraphics[width=0.238\textwidth]{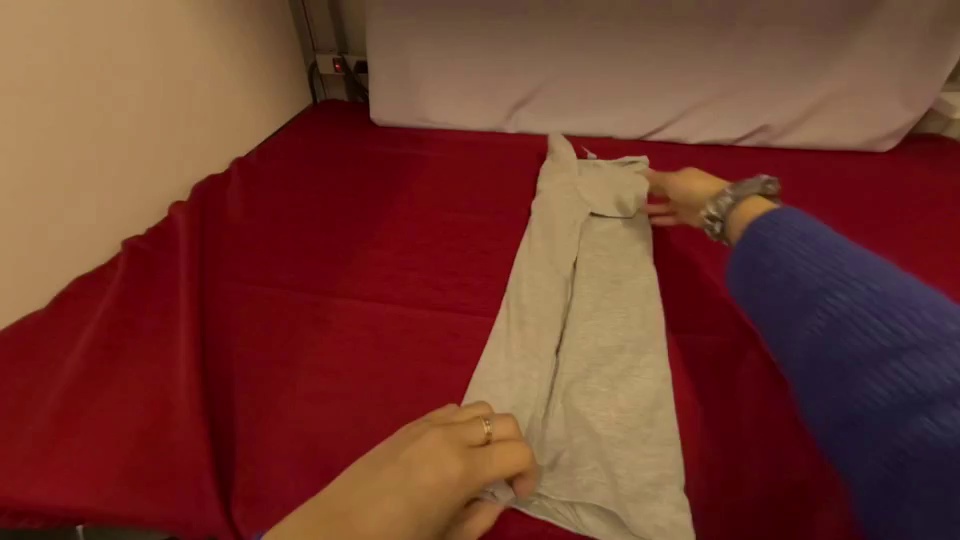}}\hfill
\fbox{\includegraphics[width=0.238\textwidth]{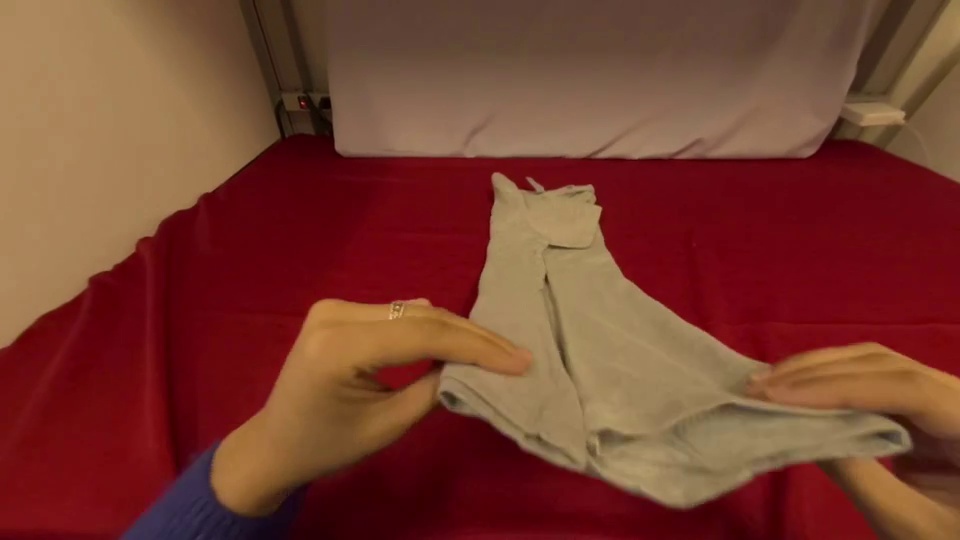}}\hfill
\fbox{\includegraphics[width=0.238\textwidth]{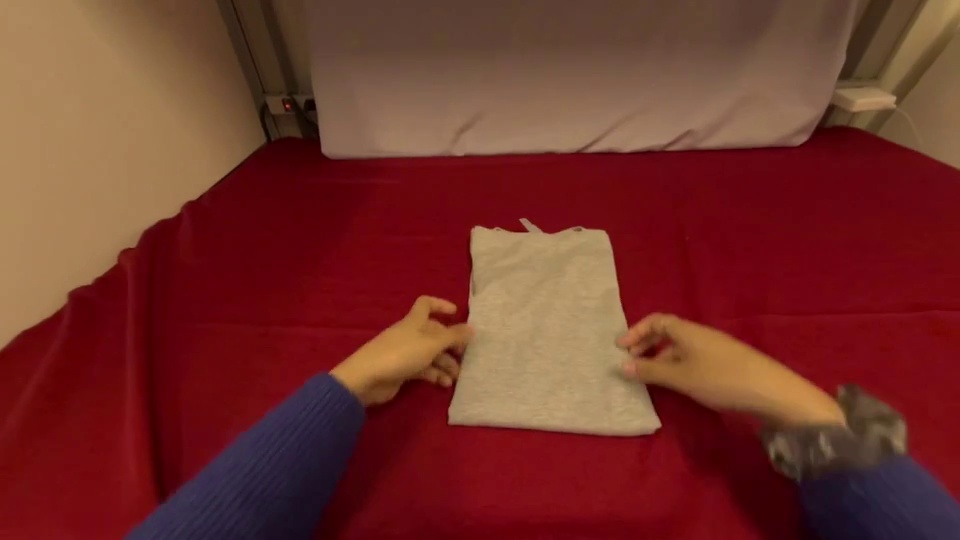}}\hfill
\fbox{\includegraphics[width=0.238\textwidth]{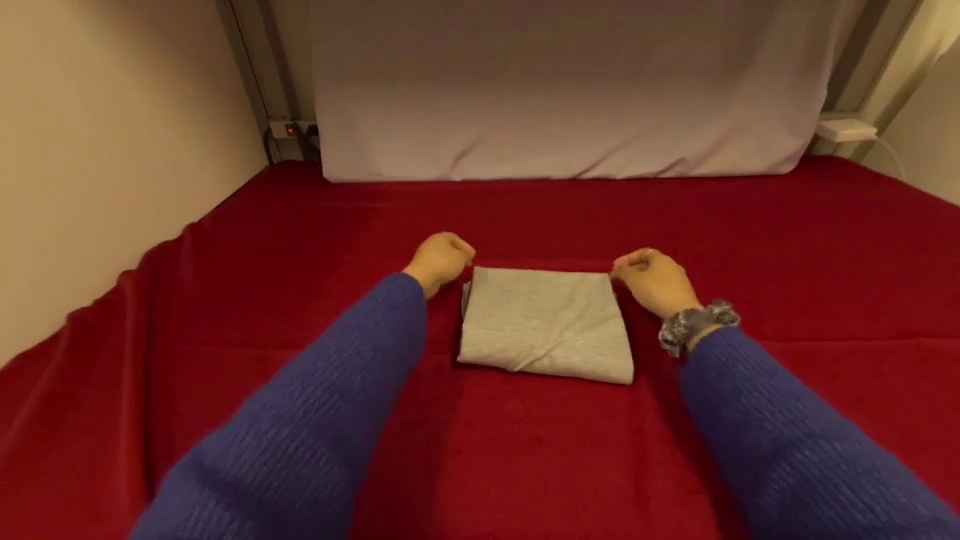}}
\par\vspace{0.35em}
\begin{tabular}{@{}p{0.30\textwidth}@{\hspace{0.02\textwidth}}p{0.68\textwidth}@{}}
\footnotesize\textbf{Caption.}
Left hand: Place the folded cloth on the red surface. Right hand: Fold the cloth in half vertically.
&
\footnotesize\textbf{Plan.}
\textbf{1.} The person places a folded piece of cloth on a red surface.
\textbf{2.} They fold the cloth in half vertically with both hands.
\textbf{3.} They then fold the cloth in half horizontally, completing the folding process.
\end{tabular}

\par\vspace{0.6em}\hrule\vspace{0.65em}
\small\textbf{(b) Scooping ice into a cup (Part~4; audit case \#43).}\par\vspace{0.25em}
\fbox{\includegraphics[width=0.238\textwidth]{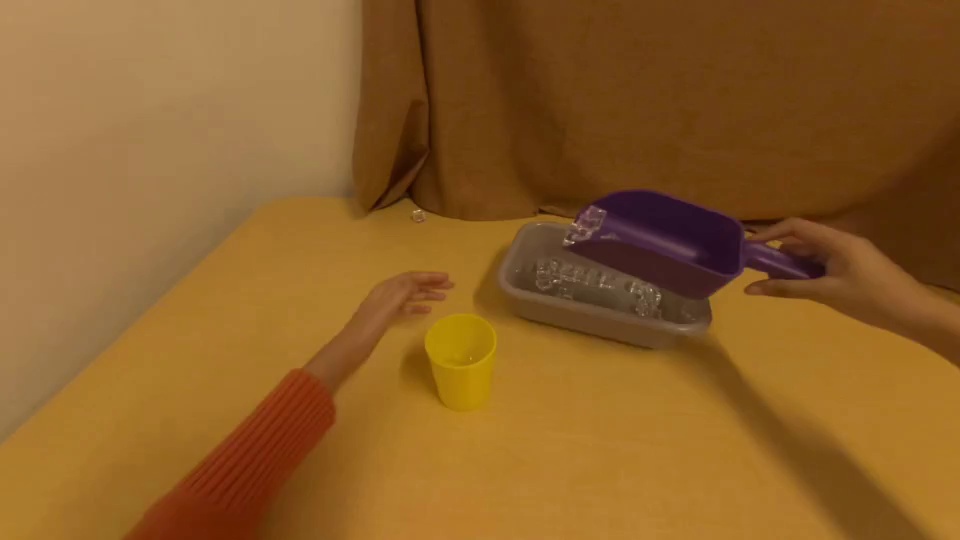}}\hfill
\fbox{\includegraphics[width=0.238\textwidth]{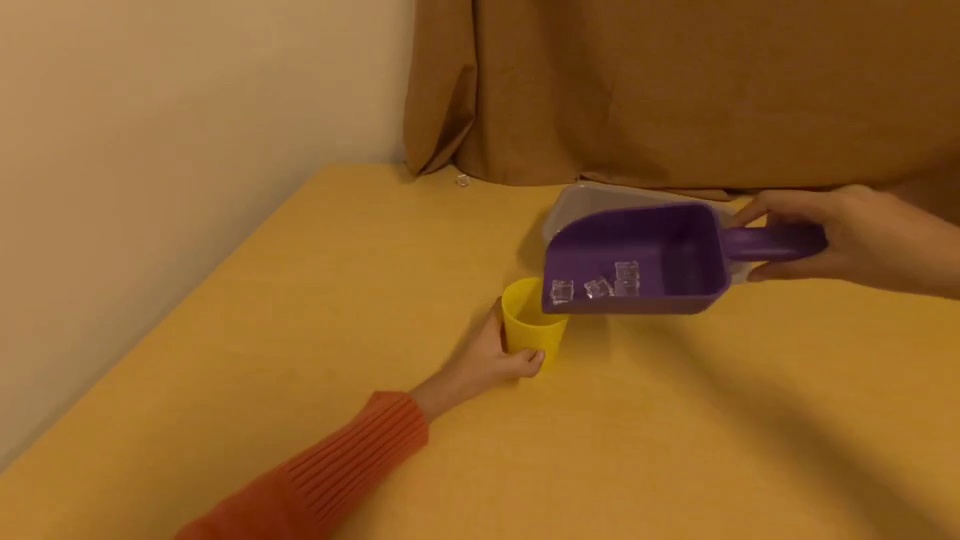}}\hfill
\fbox{\includegraphics[width=0.238\textwidth]{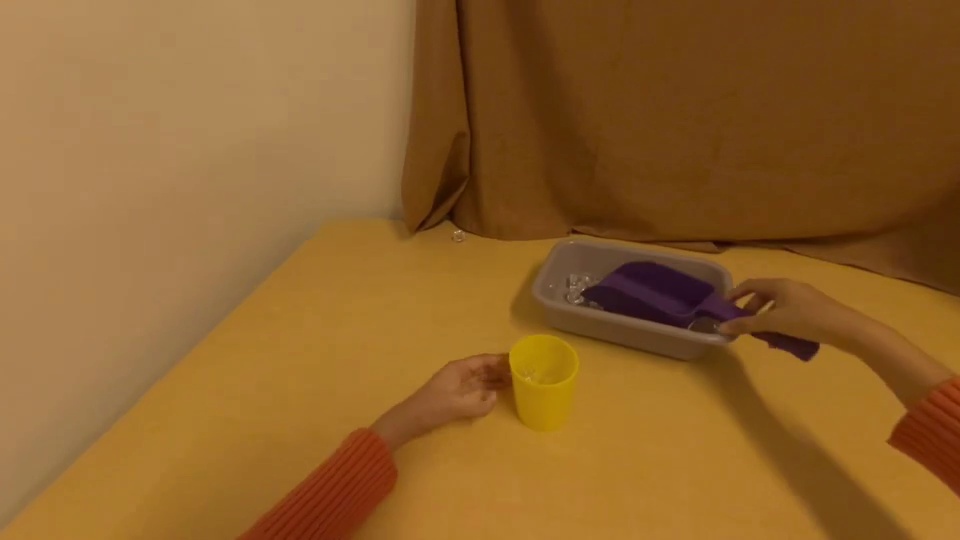}}\hfill
\fbox{\includegraphics[width=0.238\textwidth]{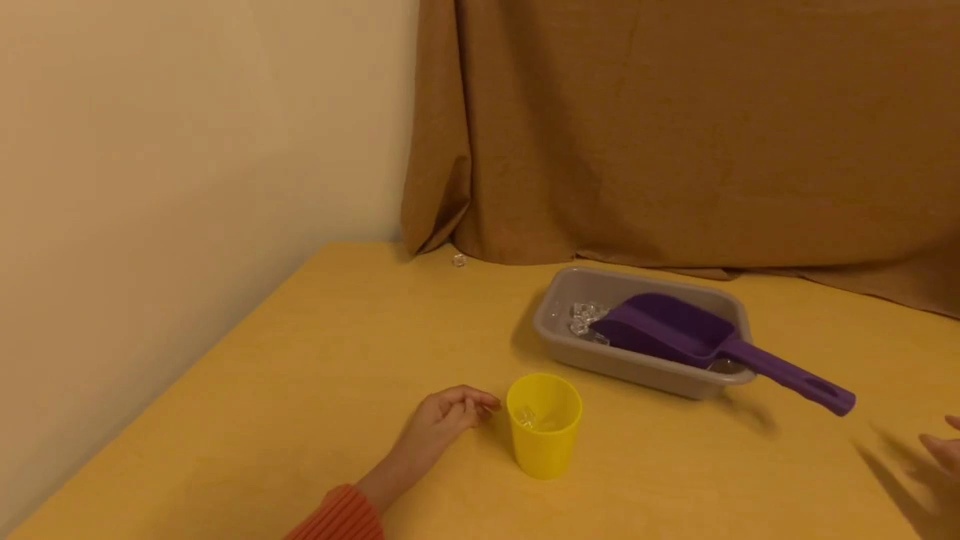}}
\par\vspace{0.35em}
\begin{tabular}{@{}p{0.30\textwidth}@{\hspace{0.02\textwidth}}p{0.68\textwidth}@{}}
\footnotesize\textbf{Caption.}
Left hand: Pick up the yellow cup. Right hand: Scoop ice cubes into the cup.
&
\footnotesize\textbf{Plan.}
\textbf{1.} The person reaches for a yellow cup with their left hand.
\textbf{2.} They pick up a purple scoop with their right hand.
\textbf{3.} The person scoops ice cubes from a container into the yellow cup.
\textbf{4.} After filling the cup, they place the scoop back into the container.
\textbf{5.} The person adjusts the position of the cup on the table.
\end{tabular}

\par\vspace{0.6em}\hrule\vspace{0.65em}
\small\textbf{(c) Connecting an AirPods charging cable (Part~1; audit case \#62).}\par\vspace{0.25em}
\fbox{\includegraphics[width=0.238\textwidth]{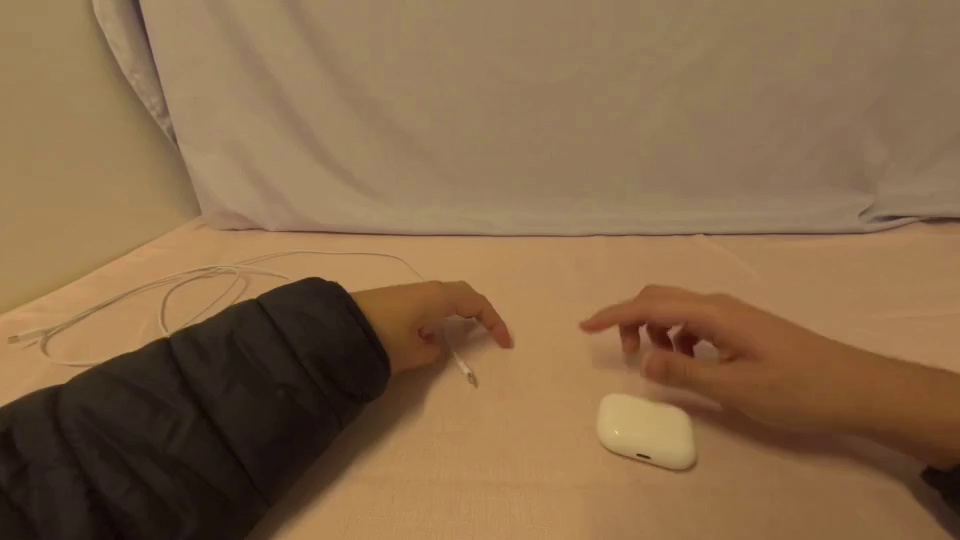}}\hfill
\fbox{\includegraphics[width=0.238\textwidth]{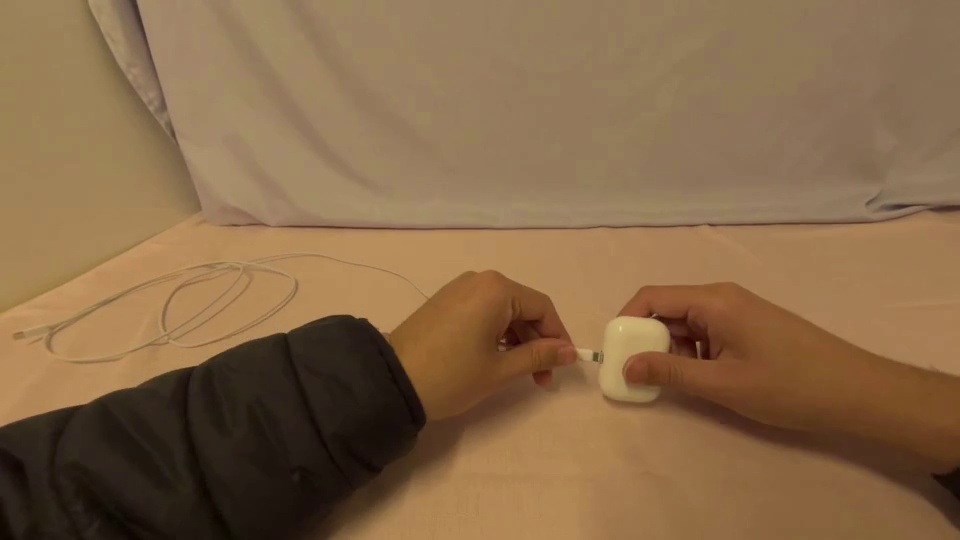}}\hfill
\fbox{\includegraphics[width=0.238\textwidth]{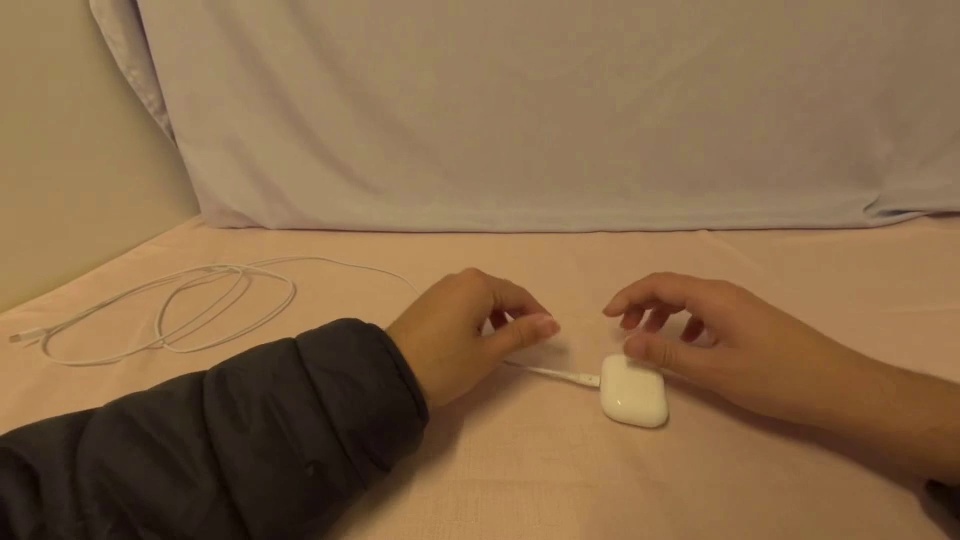}}\hfill
\fbox{\includegraphics[width=0.238\textwidth]{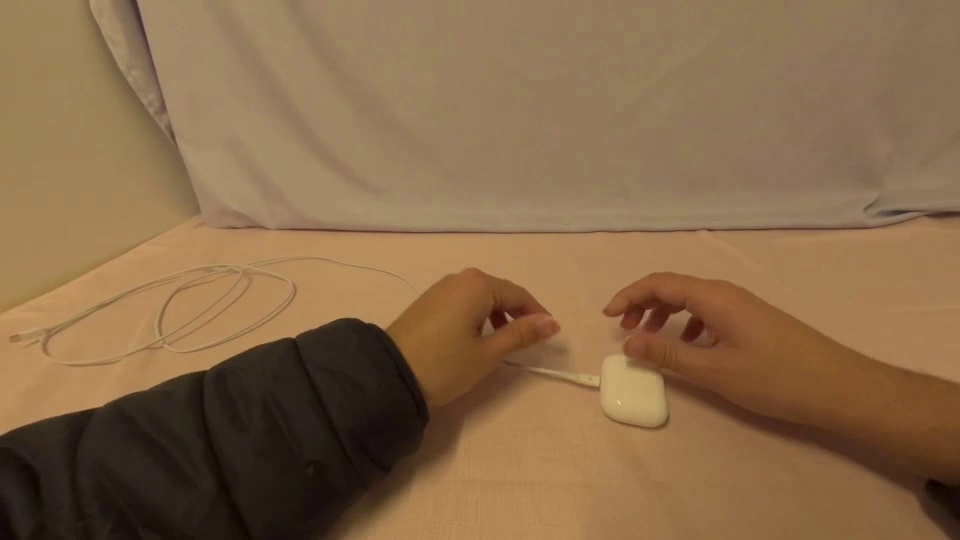}}
\par\vspace{0.35em}
\begin{tabular}{@{}p{0.30\textwidth}@{\hspace{0.02\textwidth}}p{0.68\textwidth}@{}}
\footnotesize\textbf{Caption.}
Left hand: Insert the USB cable into the white device. Right hand: Hold the white device steady.
&
\footnotesize\textbf{Plan.}
\textbf{1.} The left hand picks up the USB cable from the bed.
\textbf{2.} The left hand inserts the USB cable into the white device held by the right hand.
\textbf{3.} The right hand holds the white device steady while the left hand completes the insertion of the USB cable.
\end{tabular}

\endgroup
\caption{\textbf{Some annotation examples from the human audit.} Each row shows four uniformly spaced ground-truth frames from an audited five-second window, followed by its caption and temporally ordered motion-plan steps.}
\label{fig:audit-fullscore}
\end{figure*}

\paragraph{Qualitative examples.}
Figure~\ref{fig:audit-fullscore} visualizes three cases sampled from distinct training partitions. The ordered ground-truth frames demonstrate that the captions correctly identify the visible hand roles and manipulation task, while the corresponding motion plan decomposes the same interaction into temporally ordered and semantically consistent steps.

\begin{table*}[t]
\centering
\small
\setlength{\tabcolsep}{3.5pt}
\renewcommand{\arraystretch}{1.08}
\begin{tabular}{
>{\raggedright\arraybackslash}p{0.20\textwidth}
>{\raggedright\arraybackslash}p{0.37\textwidth}
>{\raggedright\arraybackslash}p{0.37\textwidth}}
\toprule
Configuration & Stage~I: Learn to Plan & Stage~II: Learn to Act \\
\midrule
Trainable modules
& Gemma-2 decoder, multimodal projector, depth mapper, and FoV encoder; SigLIP and DA3 encoders are frozen
& DiT-B and conditioning projectors; the Planner and all condition encoders are frozen \\
Training objective
& Autoregressive plan-language cross-entropy
& Flow matching with $t\!\sim\!\mathrm{Beta}(1.5,1.0)$ on $[0,1]$ \\
Training length
& $1$ epoch ($10{,}171$ optimizer steps)
& $4$ epochs ($40{,}684$ optimizer steps) \\
Optimizer
& AdamW, $(\beta_1,\beta_2){=}(0.9,0.999)$, weight decay $0.1$
& Same as Stage~I \\
Learning rate
& $1{\times}10^{-5}$
& $1{\times}10^{-4}$ \\
Schedule / clipping
& Constant learning rate, no warmup; gradient-norm clipping at $1.0$
& Same as Stage~I \\
Batch size
& Global batch $64$; per-device batch $2$; $4$ gradient-accumulation steps
& Global batch $64$; per-device batch $4$; $2$ gradient-accumulation steps \\
Compute
& $8{\times}$NVIDIA A40; bfloat16; FSDP full sharding with activation checkpointing
& Same as Stage~I \\
\bottomrule
\end{tabular}
\caption{\textbf{Training details.} Stage-specific settings from the final Stage~I and Stage~II checkpoint configurations.}
\label{tab:training-config}
\end{table*}

\section{Training and Evaluation}

\subsection{Evaluation Protocol}
\label{sec:eval-protocol}
All models are evaluated on the same EgoDex test set, which contains $6{,}836$ five-second forecasting windows covering all $111$ tasks. Given each observation window, the model predicts $60$ future frames at $12$~FPS for both hands. For each test window, we generate $K{=}8$ stochastic predictions and recover the corresponding joint trajectories for evaluation. We report the \emph{best-of-8} performance by selecting the prediction with the lowest overall MPJPE. The MPJPE is computed as the mean Euclidean distance over valid frames and joints across both hands. Wrist error is measured using joint $0$, while finger-relative error is computed after subtracting the wrist position from all joints before comparison. All errors are reported in millimeters in the absolute camera coordinate frame and averaged across all test windows.

\subsection{Baseline Implementations}
\noindent\textbf{VITRA re-implementation.} 
We adopt a re-implementation of VITRA \citep{vitra} as the no-plan baseline. Since the released VITRA model is trained under a different data configuration and predicts short action chunks rather than five-second future trajectories, we re-train the model under our experimental setting for a fair comparison.
Our implementation follows the same architecture and training configuration as our Stage~II model, including the PaliGemma-2-3B backbone, DiT action expert, EMPIRE-651K training data, four-epoch schedule, and flow-matching objective. This baseline is trained in a single stage directly from coarse captions, without explicit motion plans or monocular depth inputs, and jointly optimizes the VLM and DiT modules.

\noindent\textbf{Being-H0 adaptation.} We evaluate the released Being-H0 models (1B, 8B, 14B) \citep{beingh0} using their official inference protocol.  Motion generation follows the original sampling procedure with fixed block-length and block-count control. The input consists of the current EgoDex RGB frame, warped from the original camera intrinsics to Being-H0's canonical camera space, and the same caption instruction used by our models. Since Being-H0 does not use the current hand state as input, no pose initialization is provided.

We generate $5$ seconds of motion generation, (75 frames at 15 FPS). The generated wrist and finger tokens are decoded  into camera-frame MANO trajectories using the GRVQ-8K tokenizer, and linearly resampled to our 60-frame, 12~FPS evaluation protocol. A same-pose alignment test verifies the consistency between Being-H0 and our MANO joint conventions, with an average residual of approximately 5~mm caused by different fingertip definitions.


To compensate for the lack of initial hand state, we additionally evaluate a wrist-anchored variant by aligning the first-frame wrist position with the ground truth. This improves Being-H0-1B from 115.61~mm to 96.55~mm, Being-H0-8B from 85.39~mm to 81.22~mm, and Being-H0-14B from 84.90~mm to 81.21~mm. While larger Being-H0 models benefit from improved global wrist localization, our lower finger-relative error demonstrates the advantage of explicit per-hand plans for fine-grained long-horizon dexterous motion forecasting.

\begin{table*}[t]
\centering
\setlength{\tabcolsep}{6pt}
\begin{tabular}{l l c c c c}
\toprule
Run & Trainable & Epoch & Steps & s/step & Hours \\
\midrule
Baseline (single stage) & VLM $+$ DiT & $4$ & $40{,}684$ & $10.26$ & $116$ \\
\midrule
Ours Stage-I (plan SFT) & VLM & $1$ & $10{,}171$ & $12.14$ & $34$ \\
Ours Stage-II (DiT) & DiT & $4$ & $40{,}684$ & $3.23$ & $37$ \\
Ours total & --- & $5$ & $50{,}855$ & --- & $\mathbf{71}$ \\
\bottomrule
\end{tabular}
\caption{\textbf{Gradient-optimization cost on EMPIRE-651K.}} 
\label{tab:training-cost}
\end{table*}

\subsection{Training Cost}
\label{sec:training-cost}
The proposed two-stage training strategy reduces optimization cost compared with the single-stage baseline. Table~\ref{tab:training-cost} reports the gradient-optimization time over all five EMPIRE-651K partitions. All runs use the same hardware setup (8 A40 GPUs) and batch size (64 windows per optimizer step). The offline generation of Stage-I predicted plans is treated as preprocessing and excluded.

The single-stage baseline jointly fine-tunes the 3B VLM and DiT for four epochs, requiring 40,684 steps at 10.26 s/step and 116 hours in total. In contrast, our method separates plan learning and motion generation. Stage-I updates only the 3B VLM for one epoch (10,171 steps, 34 hours), while Stage-II freezes the VLM and trains the 182M DiT for four epochs (40,684 steps, 37 hours). As a result, our complete training pipeline requires 71 hours, reducing the optimization cost by 39\% compared with the baseline, despite using one additional training epoch.

The efficiency gain comes from decoupling semantic learning from motion generation: the expensive VLM backbone is optimized only during plan learning, while motion synthesis training updates only the lightweight DiT.

\subsection{Inference Cost}
\label{sec:inference-cost}
Table~\ref{tab:inference} reports the end-to-end wall-clock time per test window under the same protocol used for the main-paper results. For each window, a single GPU generates the complete best-of-$8$ prediction, including preprocessing, model inference, and trajectory decoding. As shown in Table~\ref{tab:inference}, our  model requires only $1.0$~s per window,  substantially faster than Being-H0 variants ($15.6$--$83.5$~s). 

The efficiency advantage comes from the different generation paradigms. Our model generates a compact manipulation plan with at most $96$ autoregressive tokens, whose hidden states are reused to condition all $8$ DiT samples. The VLM is therefore executed only once, while motion synthesis is performed through horizon-parallel flow matching with $4$ Euler steps. In contrast, Being-H0 directly generates motion tokens autoregressively, requiring approximately $10{,}400$ sequential token generations under the best-of-$8$ setting. Since each token depends on previous outputs, its inference cost scales with both the forecasting horizon and the number of samples.

The caption-only baseline requires only $0.4$~s per window because it entirely removes autoregressive plan generation. Compared with this baseline, our additional cost is limited to generating a short manipulation plan, while avoiding the expensive autoregressive motion synthesis. These results demonstrate that explicit planning provides an effective balance between semantic guidance and inference efficiency.

\begin{table}[t]
\centering
\small
\setlength{\tabcolsep}{5pt}
\begin{tabular}{l c c}
\toprule
Model & AR tokens & s\,/\,window \\
\midrule
Baseline (no plan) & $0$ & $0.4$ \\
\methodname{} (plan $+$ depth) & ${\le}96$ & $\mathbf{1.0}$ \\
\midrule
Being-H0-1B & ${\sim}10{,}400$ & $15.6$ \\
Being-H0-8B & ${\sim}10{,}400$ & $28.3$ \\
Being-H0-14B & ${\sim}10{,}400$ & $83.5$ \\
\bottomrule
\end{tabular}
\caption{\textbf{End-to-end inference cost.}} 
\label{tab:inference}
\end{table}

\section{Ablation Analysis}
\subsection{Stage~II Training on Predicted Motion Plans}
\label{sec:genplan}
During deployment, Stage~II is conditioned on motion plans generated by the frozen Stage~I planner. Training the motion generator with ground-truth plans would introduce a train--test mismatch by providing cleaner conditioning signals than those available during inference. Table~\ref{tab:genplan} compares training Stage~II with ground-truth plans and with plans predicted by Stage~I. For predicted-plan training, Stage~I generates motion plans for training windows using the same inference setting, and Stage~II is trained with these predicted plans as conditioning inputs. Since Stage~I remains frozen, this only changes the conditioning data without introducing additional optimization cost. Table~\ref{tab:genplan} shows that predicted-plan training consistently improves performance. On the Part~1 $26$-task subset, MPJPE decreases from $94.67$ to $93.40$~mm. When trained on all $111$ tasks, the improvement increases from $90.22$ to $84.53$~mm. These results demonstrate that predicted-plan training better matches the inference-time conditioning distribution, and we therefore adopt Stage-I predicted plans for Stage~II in the final model.

\begin{table}[h]
\centering
\small
\setlength{\tabcolsep}{5pt}
\begin{tabular}{l cc}
\toprule
Stage-II plan condition & 26-task subset & Train Part~1--5 \\
\midrule
Ground-truth plan & $94.67$ & $90.22$ \\
Stage-I predicted plan (ours) & $\mathbf{93.40}$ & $\mathbf{84.53}$ \\
\bottomrule
\end{tabular}
\caption{\textbf{Stage~II training with predicted versus ground-truth plans.}} 
\label{tab:genplan}
\end{table}

\subsection{Detailed Analysis of When Motion Planning Helps?}
Table 3 of the main paper compares the deployable model with self-generated-plan against the no-plan baseline on the EgoDex test set. As shown, explicit motion planning improves forecasting performance on average, but the benefit is not uniform across tasks. 

Let $\Delta$ denote the per-task MPJPE difference between the planning model and the baseline (negative values indicate improvement). As shown in Table 3 of the main paper, the self-generated plan improves performance on $89/111$ tasks ($80\%$), achieving a sample-weighted average gain of $22.3$~mm and a median per-task improvement of $14.3$~mm. Moreover, the benefit increases with task difficulty: baseline MPJPE and $\Delta$ exhibit a Pearson correlation of $-0.567$, indicating larger improvements for more challenging tasks. Consistent with the difficulty analysis in the main paper, planning provides limited gains on easy tasks ($\Delta=-4.6$~mm, improving $25/43$ tasks), but substantially larger improvements on medium ($-26.2$~mm, $47/49$ tasks) and hard tasks ($-49.7$~mm, $17/19$ tasks).

Planning also provides stronger benefits for unseen tasks. Compared with the Part~1 training setting, the self-generated plan improves $15/26$ seen tasks ($58\%$, weighted $\Delta=-10.9$~mm) and $74/85$ unseen tasks ($87\%$, weighted $\Delta=-24.8$~mm), yielding more than twice the improvement on unseen scenarios. The largest gains are observed on complex, long-horizon manipulation tasks, including \texttt{wash\_fruit} ($343{\to}154$~mm), \texttt{stock\_unstock\_fridge} ($303{\to}185$~mm), \texttt{wash\_put\_away\_dishes} ($234{\to}126$~mm), and \texttt{flip\_pages} ($164{\to}58$~mm). Performance degradation is relatively rare and is primarily observed in two scenarios: simple tasks with low baseline errors, where additional planning constraints may limit motion flexibility, and highly challenging tasks where the predicted plans may still contain inaccuracies.


\subsection{Qualitative Hand-Motion Forecasts}
\label{sec:qualitative-motion-forecasts}

\begin{figure}[h]
\centering
\includegraphics[width=\columnwidth]{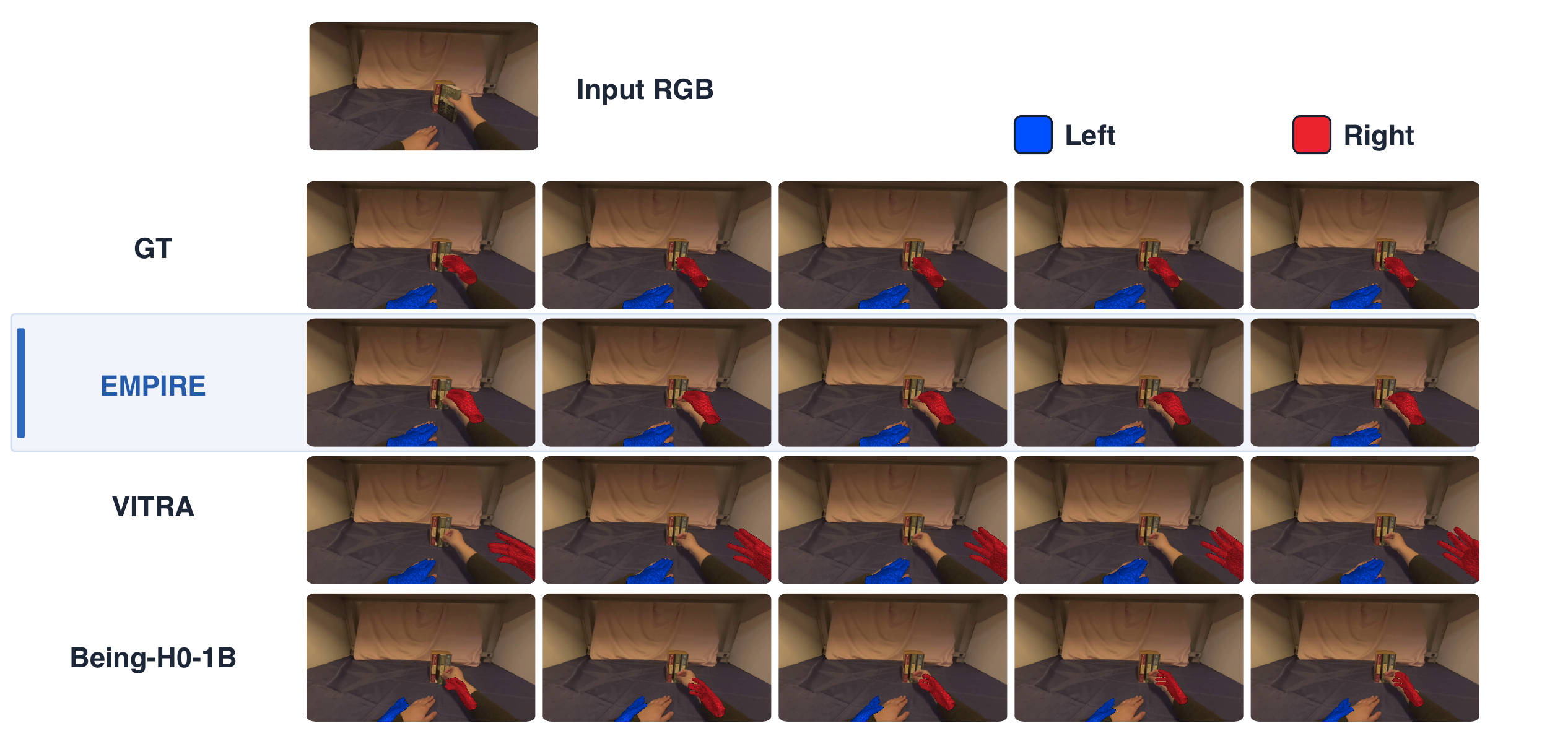}
\caption{Qualitative comparison of motion predictions
against baselines. \texttt{Task: insert\_remove\_bookshelf}.}
\label{fig:qual-motion-case12}
\end{figure}

\begin{figure}[h]
\centering
\includegraphics[width=\columnwidth]{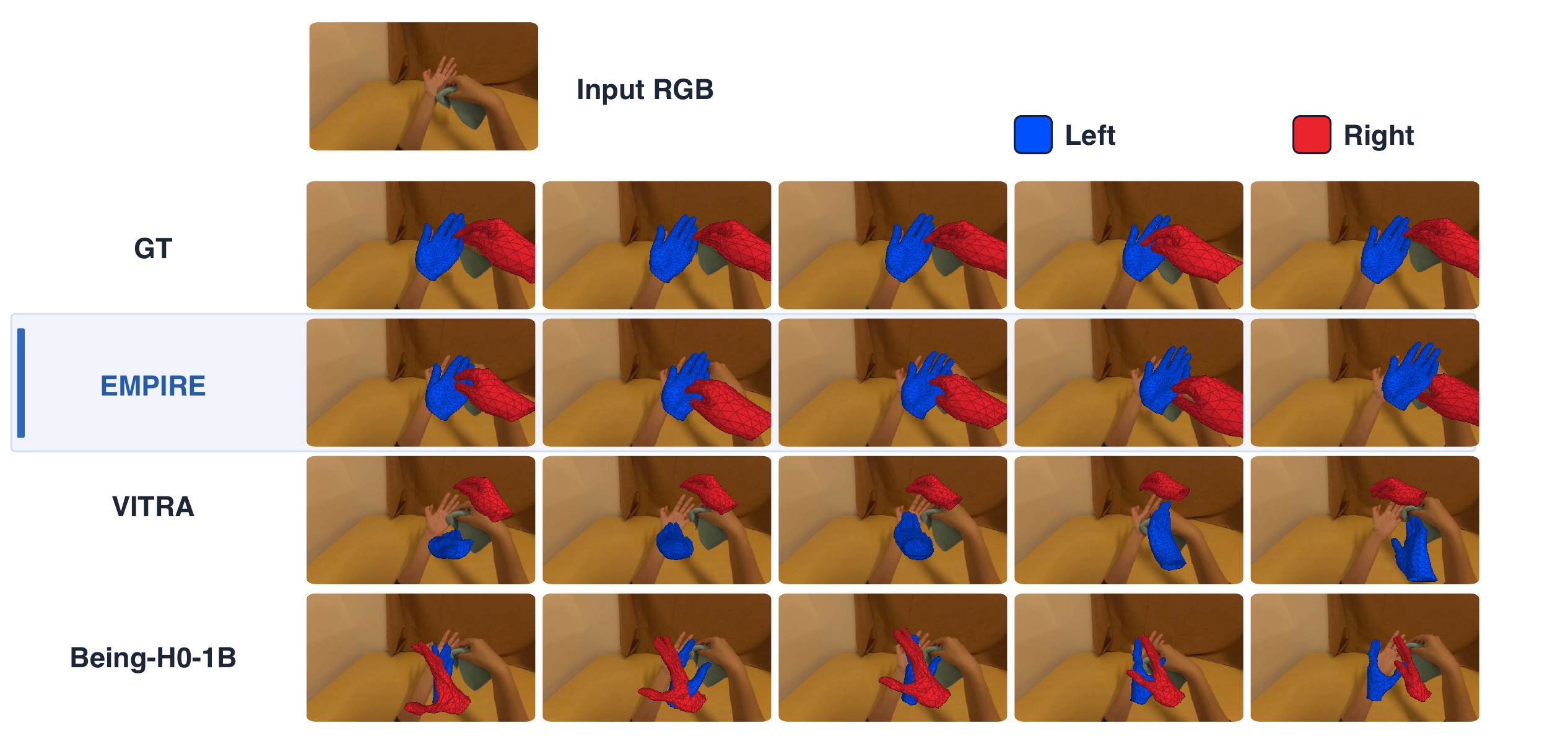}
\caption{Qualitative comparison of motion predictions
against baselines. \texttt{Task: dry\_hands}.}
\label{fig:qual-motion-case13}
\end{figure}

Figures~\ref{fig:qual-motion-case12} and~\ref{fig:qual-motion-case13} present qualitative comparisons of future hand-motion forecasts on two representative tasks. All predictions are rendered on the same five future RGB frames sampled from the $60$-frame, $12$~fps forecasting horizon. The ground-truth row uses the corresponding EgoDex MANO annotations, and all predictions are transformed into the corresponding future camera views before rendering. Blue and red denote the left and right hands, respectively.

As shown in the figures, EMPIRE better captures the temporal evolution of bimanual manipulation compared with existing baselines. In particular, it preserves more accurate hand-object interactions and finger articulations over long horizons, while VITRA and Being-H0 exhibit larger deviations in hand placement and coordination. These results demonstrate that explicit manipulation plans provide effective guidance for long-horizon hand-motion forecasting.

%% file: paper_refs.bib
@misc{vitra,
  title={Scalable Vision-Language-Action Model Pretraining for Robotic Manipulation with Real-Life Human Activity Videos},
  author={Li, Qixiu and Deng, Yu and Liang, Yaobo and Luo, Lin and Zhou, Lei and Yao, Chengtang and Zeng, Lingqi and Feng, Zhiyuan and Liang, Huizhi and Xu, Sicheng and Zhang, Yizhong and Chen, Xi and Chen, Hao and Sun, Lily and Chen, Dong and Yang, Jiaolong and Guo, Baining},
  year={2025},
  eprint={2510.21571},
  archivePrefix={arXiv},
  primaryClass={cs.RO},
  url={https://arxiv.org/abs/2510.21571},
}

@misc{beingh0,
  title={Being-H0: Vision-Language-Action Pretraining from Large-Scale Human Videos},
  author={Luo, Hao and Feng, Yicheng and Zhang, Wanpeng and Zheng, Sipeng and Wang, Ye and Yuan, Haoqi and Liu, Jiazheng and Xu, Chaoyi and Jin, Qin and Lu, Zongqing},
  year={2025},
  eprint={2507.15597},
  archivePrefix={arXiv},
  primaryClass={cs.RO},
  url={https://arxiv.org/abs/2507.15597},
}

@inproceedings{lipman2023flow,
  title={Flow Matching for Generative Modeling},
  author={Lipman, Yaron and Chen, Ricky T. Q. and Ben-Hamu, Heli and Nickel, Maximilian and Le, Matt},
  booktitle={International Conference on Learning Representations (ICLR)},
  year={2023},
  url={https://openreview.net/forum?id=PqvMRDCJT9t},
}

@inproceedings{peebles2023dit,
  title={Scalable Diffusion Models with Transformers},
  author={Peebles, William and Xie, Saining},
  booktitle={Proceedings of the IEEE/CVF International Conference on Computer Vision (ICCV)},
  pages={4195--4205},
  month={October},
  year={2023},
  url={https://openaccess.thecvf.com/content/ICCV2023/html/Peebles_Scalable_Diffusion_Models_with_Transformers_ICCV_2023_paper.html},
}

@misc{paligemma2,
  title={PaliGemma 2: A Family of Versatile VLMs for Transfer},
  author={Steiner, Andreas and Susano Pinto, Andr\'e and Tschannen, Michael and Keysers, Daniel and Wang, Xiao and Bitton, Yonatan and Gritsenko, Alexey and Minderer, Matthias and Sherbondy, Anthony and Long, Shangbang and Qin, Siyang and Ingle, Reeve and Bugliarello, Emanuele and Kazemzadeh, Sahar and Mesnard, Thomas and Alabdulmohsin, Ibrahim and Beyer, Lucas and Zhai, Xiaohua},
  year={2024},
  eprint={2412.03555},
  archivePrefix={arXiv},
  primaryClass={cs.CV},
  url={https://arxiv.org/abs/2412.03555},
}

@inproceedings{rt2,
  title={RT-2: Vision-Language-Action Models Transfer Web Knowledge to Robotic Control},
  author={Brohan, Anthony and Brown, Noah and Carbajal, Justice and others},
  booktitle={Proceedings of The 7th Conference on Robot Learning},
  pages={2165--2183},
  year={2023},
  volume={229},
  series={Proceedings of Machine Learning Research},
  publisher={PMLR},
  url={https://proceedings.mlr.press/v229/zitkovich23a.html},
}

@inproceedings{ecot,
  title={Robotic Control via Embodied Chain-of-Thought Reasoning},
  author={Zawalski, Micha\l{} and Chen, William and Pertsch, Karl and Mees, Oier and Finn, Chelsea and Levine, Sergey},
  booktitle={Proceedings of The 8th Conference on Robot Learning},
  pages={3157--3181},
  year={2025},
  volume={270},
  series={Proceedings of Machine Learning Research},
  publisher={PMLR},
  url={https://proceedings.mlr.press/v270/zawalski25a.html},
}

@article{mano,
  title={Embodied Hands: Modeling and Capturing Hands and Bodies Together},
  author={Romero, Javier and Tzionas, Dimitrios and Black, Michael J.},
  journal={ACM Transactions on Graphics (ToG)},
  volume={36},
  number={6},
  year={2017},
  publisher={ACM},
}

@inproceedings{egodex,
  title={EgoDex: Learning Dexterous Manipulation from Large-Scale Egocentric Video},
  author={Hoque, Ryan and Huang, Peide and Yoon, David J. and Sivapurapu, Mouli and Zhang, Jian},
  booktitle={International Conference on Learning Representations (ICLR)},
  year={2026},
  url={https://openreview.net/forum?id=FFxkFMU89E},
}

@misc{qwen3,
  title={Qwen3 Technical Report},
  author={Yang, An and Li, Anfeng and Yang, Baosong and others},
  year={2025},
  eprint={2505.09388},
  archivePrefix={arXiv},
  primaryClass={cs.CL},
  url={https://arxiv.org/abs/2505.09388},
}

@inproceedings{pi0,
  title={{$\pi_0$}: A Vision-Language-Action Flow Model for General Robot Control},
  author={Black, Kevin and Brown, Noah and Driess, Danny and Esmail, Adnan and Equi, Michael Robert and Finn, Chelsea and Fusai, Niccolo and Groom, Lachy and Hausman, Karol and Ichter, Brian and Jakubczak, Szymon and Jones, Tim and Ke, Liyiming and Levine, Sergey and Li-Bell, Adrian and Mothukuri, Mohith and Nair, Suraj and Pertsch, Karl and Shi, Lucy Xiaoyang and Smith, Laura and Tanner, James and Vuong, Quan and Walling, Anna and Wang, Haohuan and Zhilinsky, Ury},
  booktitle={Proceedings of Robotics: Science and Systems},
  year={2025},
  month={June},
  doi={10.15607/RSS.2025.XXI.010},
  url={https://www.roboticsproceedings.org/rss21/p010.html},
}

@inproceedings{ego4d,
  title={Ego4D: Around the World in 3,000 Hours of Egocentric Video},
  author={Grauman, Kristen and Westbury, Andrew and Byrne, Eugene and Chavis, Zachary and Furnari, Antonino and Girdhar, Rohit and Hamburger, Jackson and Jiang, Hao and others},
  booktitle={Proceedings of the IEEE/CVF Conference on Computer Vision and Pattern Recognition (CVPR)},
  pages={18995--19012},
  month={June},
  year={2022},
  url={https://openaccess.thecvf.com/content/CVPR2022/html/Grauman_Ego4D_Around_the_World_in_3000_Hours_of_Egocentric_Video_CVPR_2022_paper.html},
}

@inproceedings{siglip,
  title={Sigmoid Loss for Language Image Pre-Training},
  author={Zhai, Xiaohua and Mustafa, Basil and Kolesnikov, Alexander and Beyer, Lucas},
  booktitle={Proceedings of the IEEE/CVF International Conference on Computer Vision (ICCV)},
  pages={11975--11986},
  month={October},
  year={2023},
  url={https://openaccess.thecvf.com/content/ICCV2023/html/Zhai_Sigmoid_Loss_for_Language_Image_Pre-Training_ICCV_2023_paper.html},
}

@inproceedings{llava,
  title={Visual Instruction Tuning},
  author={Liu, Haotian and Li, Chunyuan and Wu, Qingyang and Lee, Yong Jae},
  booktitle={Advances in Neural Information Processing Systems},
  volume={36},
  pages={34892--34916},
  publisher={Curran Associates, Inc.},
  year={2023},
  doi={10.52202/075280-1516},
  url={https://proceedings.neurips.cc/paper_files/paper/2023/hash/6dcf277ea32ce3288914faf369fe6de0-Abstract-Conference.html},
}

@inproceedings{openvla,
  title={OpenVLA: An Open-Source Vision-Language-Action Model},
  author={Kim, Moo Jin and Pertsch, Karl and Karamcheti, Siddharth and Xiao, Ted and Balakrishna, Ashwin and Nair, Suraj and Rafailov, Rafael and Foster, Ethan P. and Sanketi, Pannag R. and Vuong, Quan and Kollar, Thomas and Burchfiel, Benjamin and Tedrake, Russ and Sadigh, Dorsa and Levine, Sergey and Liang, Percy and Finn, Chelsea},
  booktitle={Proceedings of The 8th Conference on Robot Learning},
  pages={2679--2713},
  year={2025},
  volume={270},
  series={Proceedings of Machine Learning Research},
  publisher={PMLR},
  url={https://proceedings.mlr.press/v270/kim25c.html},
}

@inproceedings{depthanything3,
  title={Depth Anything 3: Recovering the Visual Space from Any Views},
  author={Lin, Haotong and Chen, Sili and Liew, Jun Hao and Chen, Donny Y. and Li, Zhenyu and Zhao, Yang and Peng, Sida and Guo, Hengkai and Zhou, Xiaowei and Shi, Guang and Feng, Jiashi and Kang, Bingyi},
  booktitle={International Conference on Learning Representations (ICLR)},
  year={2026},
  url={https://openreview.net/forum?id=yirunib8l8},
}

@inproceedings{kingma2015adam,
  title={Adam: A Method for Stochastic Optimization},
  author={Kingma, Diederik P. and Ba, Jimmy},
  booktitle={International Conference on Learning Representations (ICLR)},
  year={2015},
}

@misc{qwen25vl,
  title={Qwen2.5-VL Technical Report},
  author={Bai, Shuai and Chen, Keqin and Liu, Xuejing and Wang, Jialin and Ge, Wenbin and Song, Sibo and Dang, Kai and Wang, Peng and Wang, Shijie and Tang, Jun and others},
  year={2025},
  eprint={2502.13923},
  archivePrefix={arXiv},
  primaryClass={cs.CV},
  url={https://arxiv.org/abs/2502.13923},
}

@misc{qwen3vl,
  title={{Qwen3-VL} Technical Report},
  author={Bai, Shuai and Cai, Yuxuan and Chen, Ruizhe and others},
  year={2025},
  eprint={2511.21631},
  archivePrefix={arXiv},
  primaryClass={cs.CV},
  url={https://arxiv.org/abs/2511.21631},
}

@inproceedings{spatialvlm,
  title={{SpatialVLM}: Endowing Vision-Language Models with Spatial Reasoning Capabilities},
  author={Chen, Boyuan and Xu, Zhuo and Kirmani, Sean and Ichter, Brian and Sadigh, Dorsa and Guibas, Leonidas and Xia, Fei},
  booktitle={Proceedings of the IEEE/CVF Conference on Computer Vision and Pattern Recognition (CVPR)},
  pages={14455--14465},
  year={2024},
}

@inproceedings{vsibench,
  title={Thinking in Space: How Multimodal Large Language Models See, Remember, and Recall Spaces},
  author={Yang, Jihan and Yang, Shusheng and Gupta, Anjali W. and Han, Rilyn and Fei-Fei, Li and Xie, Saining},
  booktitle={Proceedings of the IEEE/CVF Conference on Computer Vision and Pattern Recognition (CVPR)},
  pages={10632--10643},
  year={2025},
}

@inproceedings{spatial457,
  title={{Spatial457}: A Diagnostic Benchmark for {6D} Spatial Reasoning of Large Multimodal Models},
  author={Wang, Xingrui and Ma, Wufei and Zhang, Tiezheng and de Melo, Celso M. and Chen, Jieneng and Yuille, Alan},
  booktitle={Proceedings of the IEEE/CVF Conference on Computer Vision and Pattern Recognition (CVPR)},
  pages={24669--24679},
  year={2025},
}

@inproceedings{grab,
  title     = {{GRAB}: A Dataset of Whole-Body Human Grasping of Objects},
  author    = {Taheri, Omid and Ghorbani, Nima and Black, Michael J. and Tzionas, Dimitrios},
  booktitle = {European Conference on Computer Vision (ECCV)},
  year      = {2020}
}

@inproceedings{goal,
    title={GOAL: Generating 4D Whole-Body Motion for Hand-Object Grasping},
  author={Taheri, Omid and Choutas, Vasileios and Black, Michael J and Tzionas, Dimitrios},
  booktitle={Proceedings of the IEEE/CVF Conference on Computer Vision and Pattern Recognition (CVPR)},
  pages={13253--13263},
  year={2022},
}

@article{imos,
  title={IMoS: Intent-Driven Full-Body Motion Synthesis for Human-Object Interactions},
  author={Ghosh, Anindita and Dabral, Rishabh and Golyanik, Vladislav and Theobalt, Christian and Slusallek, Philipp},
  booktitle={COMPUTER GRAPHICS forum},
  volume={42},
  number={2},
  year={2023}
}

@inproceedings{text2hoi,
  title={Text2hoi: Text-guided 3d motion generation for hand-object interaction},
  author={Cha, Junuk and Kim, Jihyeon and Yoon, Jae Shin and Baek, Seungryul},
  booktitle={Proceedings of the IEEE/CVF Conference on Computer Vision and Pattern Recognition (CVPR)},
  pages={1577--1585},
  year={2024}
}

@inproceedings{diffh2o,
  title     = {{DiffH2O}: Diffusion-Based Synthesis of Hand-Object Interactions from Textual Descriptions},
  author    = {Christen, Sammy and Hampali, Shreyas and Sener, Fadime and Remelli, Edoardo and Hodan, Tomas and Sauser, Eric and Ma, Shugao and Tekin, Bugra},
  booktitle = {SIGGRAPH Asia Conference Papers},
  year      = {2024}
}

@inproceedings{hoigpt,
  title     = {{HOIGPT}: Learning Long-Sequence Hand-Object Interaction with Language Models},
  author    = {Huang, Mingzhen and Chu, Fu-Jen and Tekin, Bugra and Liang, Kevin J. and Ma, Haoyu and Wang, Weiyao and Chen, Xingyu and Gleize, Pierre and Xue, Hongfei and Lyu, Siwei and Kitani, Kris and Feiszli, Matt and Tang, Hao},
  booktitle = {Proceedings of the IEEE/CVF Conference on Computer Vision and Pattern Recognition (CVPR)},
  pages     = {7136--7146},
  year      = {2025}
}

@inproceedings{megohand,
  title={Megohand: Multimodal egocentric hand-object interaction motion generation},
  author={Zhou, Bohan and Zhan, Yi and Zhang, Zhongbin and Lu, Zongqing},
  journal={Advances in Neural Information Processing Systems (NeurIPS)},
  volume={38},
  pages={49464--49490},
  year={2026}
}

@inproceedings{internvl,
  title={Intern VL: Scaling up Vision Foundation Models and Aligning for Generic Visual-Linguistic Tasks},
  author={Chen, Zhe and Wu, Jiannan and Wang, Wenhai and Su, Weijie and Chen, Guo and Xing, Sen and Zhong, Muyan and Zhang, Qinglong and Zhu, Xizhou and Lu, Lewei and others},
  booktitle={Proceedings of the IEEE/CVF Conference on Computer Vision and Pattern Recognition (CVPR)},
  pages={24185--24198},
  year={2024},
}

@inproceedings{saycan,
  title={Do As I Can, Not As I Say: Grounding Language in Robotic Affordances},
  author={Ichter, Brian and Brohan, Anthony and Chebotar, Yevgen and Finn, Chelsea and Hausman, Karol and Herzog, Alexander and Ho, Daniel and Ibarz, Julian and Irpan, Alex and Jang, Eric and others},
  booktitle={Proceedings of the 6th Conference on Robot Learning},
  series={Proceedings of Machine Learning Research},
  publisher={PMLR},
  volume={205},
  pages={287--318},
  year={2023},
  url={https://proceedings.mlr.press/v205/ichter23a.html},
}

@inproceedings{palme,
  title={{PaLM-E}: An Embodied Multimodal Language Model},
  author={Driess, Danny and Xia, Fei and Sajjadi, Mehdi S. M. and Lynch, Corey and Chowdhery, Aakanksha and Ichter, Brian and Wahid, Ayzaan and Tompson, Jonathan and Vuong, Quan and Yu, Tianhe and others},
  booktitle={Proceedings of the 40th International Conference on Machine Learning (ICML)},
  series={Proceedings of Machine Learning Research},
  publisher={PMLR},
  volume={202},
  pages={8469--8488},
  year={2023},
  url={https://proceedings.mlr.press/v202/driess23a.html},
}

@inproceedings{rth,
  title={{RT-H}: Action Hierarchies Using Language},
  author={Belkhale, Suneel and Ding, Tianli and Xiao, Ted and Sermanet, Pierre and Vuong, Quan and Tompson, Jonathan and Chebotar, Yevgen and Dwibedi, Debidatta and Sadigh, Dorsa},
  booktitle={Proceedings of Robotics: Science and Systems},
  year={2024},
  doi={10.15607/RSS.2024.XX.049},
  url={https://www.roboticsproceedings.org/rss20/p049.html},
}

@inproceedings{octo,
  title={Octo: An Open-Source Generalist Robot Policy},
  author={{Octo Model Team} and Ghosh, Dibya and Walke, Homer Rich and Pertsch, Karl and Black, Kevin and Mees, Oier and Dasari, Sudeep and Hejna, Joey and Kreiman, Tobias and Xu, Charles and Luo, Jianlan and others},
  booktitle={Proceedings of Robotics: Science and Systems},
  year={2024},
  doi={10.15607/RSS.2024.XX.090},
  url={https://www.roboticsproceedings.org/rss20/p090.html},
}

@inproceedings{h2o,
  title={H2O: Two Hands Manipulating Objects for First Person Interaction Recognition},
  author={Kwon, Taein and Tekin, Bugra and St{\"u}hmer, Jan and Bogo, Federica and Pollefeys, Marc},
  booktitle={Proceedings of the IEEE/CVF International Conference on Computer Vision (ICCV)},
  pages={10138--10148},
  year={2021},
  url={https://openaccess.thecvf.com/content/ICCV2021/html/Kwon_H2O_Two_Hands_Manipulating_Objects_for_First_Person_Interaction_Recognition_ICCV_2021_paper.html},
}

@inproceedings{arctic,
  title={{ARCTIC}: A Dataset for Dexterous Bimanual Hand-Object Manipulation},
  author={Fan, Zicong and Taheri, Omid and Tzionas, Dimitrios and Kocabas, Muhammed and Kaufmann, Manuel and Black, Michael J. and Hilliges, Otmar},
  booktitle={Proceedings of the IEEE/CVF Conference on Computer Vision and Pattern Recognition (CVPR)},
  pages={12943--12954},
  year={2023},
  url={https://openaccess.thecvf.com/content/CVPR2023/html/Fan_ARCTIC_A_Dataset_for_Dexterous_Bimanual_Hand-Object_Manipulation_CVPR_2023_paper.html},
}

@inproceedings{hot3d,
  title={{HOT3D}: Hand and Object Tracking in 3D from Egocentric Multi-View Videos},
  author={Banerjee, Prithviraj and Shkodrani, Sindi and Moulon, Pierre and Hampali, Shreyas and Han, Shangchen and Zhang, Fan and Zhang, Linguang and Fountain, Jade and Miller, Edward and Basol, Selen and others},
  booktitle={Proceedings of the IEEE/CVF Conference on Computer Vision and Pattern Recognition (CVPR)},
  pages={7061--7071},
  year={2025},
  url={https://openaccess.thecvf.com/content/CVPR2025/html/Banerjee_HOT3D_Hand_and_Object_Tracking_in_3D_from_Egocentric_Multi-View_CVPR_2025_paper.html},
}

@misc{feng2026humanvideos,
  title         = {From Human Videos to Robot Manipulation: A Survey on Scalable Vision-Language-Action Learning with Human-Centric Data},
  author        = {Feng, Zhiyuan and Li, Qixiu and Liang, Huizhi and Yang, Rushuai and Shen, Yichao and Du, Zhiying and Zhang, Zhaowei and Deng, Yu and Zhao, Li and Zhao, Hao and Lu, Zongqing and Mees, Oier and Pollefeys, Marc and Yang, Jiaolong and Guo, Baining},
  year          = {2026},
  eprint        = {2606.00054},
  archivePrefix = {arXiv},
  primaryClass  = {cs.RO},
  url           = {https://arxiv.org/abs/2606.00054}
}

@InProceedings{jiang2026crosshand,
  author    = {Jiang, Guangqi and Liang, Yutong and Ye, Jianglong and Huang, Jia-Yang and Jing, Changwei and Duan, Rocky and Abbeel, Pieter and Wang, Xiaolong and Zou, Xueyan},
    title     = {Cross-Hand Latent Representation for Vision-Language-Action Models},
    booktitle = {Proceedings of the IEEE/CVF Conference on Computer Vision and Pattern Recognition (CVPR)},
    month     = {June},
    year      = {2026},
    pages     = {13496-13507}
}

@article{pearson1895,
  title   = {{VII}. Note on Regression and Inheritance in the Case of Two Parents},
  author  = {Pearson, Karl},
  journal = {Proceedings of the Royal Society of London},
  volume  = {58},
  pages   = {240--242},
  year    = {1895},
  doi     = {10.1098/rspl.1895.0041},
  url     = {https://doi.org/10.1098/rspl.1895.0041}
}
